\documentclass{article}
\usepackage{vendor/iclr2027_conference,times}
\usepackage{amsmath,amssymb}
\usepackage{graphicx}
\usepackage{booktabs}
\usepackage{array}
\usepackage{longtable}
\usepackage{calc}
\usepackage{caption}
\usepackage{float}
\usepackage{xcolor}
\usepackage{url}
\usepackage{hyperref}
\usepackage[normalem]{ulem}
\usepackage{fancyvrb}
\usepackage{fvextra}
\usepackage{framed}

\providecommand{\tightlist}{\setlength{\itemsep}{0pt}\setlength{\parskip}{0pt}}
\definecolor{shadecolor}{RGB}{248,248,248}

\DefineVerbatimEnvironment{Highlighting}{Verbatim}{commandchars=\\\{\},fontsize=\small}

\graphicspath{{./}}
\hypersetup{colorlinks=true,linkcolor=blue!50!black,citecolor=blue!50!black,urlcolor=blue!50!black}
\hypersetup{pdftitle={Interface-Induced Trajectory Censoring: When Tool-Call Interfaces Distort Evaluation and RL Experience},pdfauthor={Wenbo Wang, Di Sang}}

\title{Interface-Induced Trajectory Censoring:\\
When Tool-Call Interfaces Distort Evaluation and RL Experience}

\author{Wenbo Wang \\
\texttt{wenbwang3-c@my.cityu.edu.hk}
\And
Di Sang \\
\texttt{3120265429@bit.edu.cn}}

\iclrfinalcopy

\begin{document}
\maketitle
\lhead{Preprint}

\begin{abstract}
Tool-use evaluations typically measure calls through a serving stack, which can report zero calls even when model outputs contain call-shaped payloads. We study interface-induced trajectory censoring: a mismatch between model emissions and the interface that converts them into actions. On selected BFCL v4 subsets, changing only the serving adapter while holding model weights and other evaluation settings fixed moves scores under the official scorer from 0.00 to 0.96 on simple\_python and from 0.00 to 0.19 on multi\_turn\_base. A template--parser factorial experiment shows strong complementarity: neither component replacement alone restores parsing from the baseline, whereas their joint replacement recovers 196/200 first-request parses. Across five Qwen2.5-Coder checkpoints, classifier-positive emissions increase with model size while the mismatched interface reports no calls; a separate matched-interface Qwen3 ladder exhibits few silent emissions through 8B. Experiments on 115 $\tau$-bench retail tasks extend the mechanism to a stateful, non-code environment. In a preregistered, single-seed, 150-step 7B LoRA experiment, changing only the parser registered in verl increases tool executions from 10 to 16,844. However, neither a shared programmatic-feedback evaluation nor a post-hoc common native function-calling evaluation detects a held-out success gain from repair; rescue counts do not increase. Interface repair thus restores tool-mediated experience without a detected performance gain in this setting. These findings show that tool-use measurements depend on the model--interface configuration and motivate layer-wise observability and conformance checks before attributing evaluation or training outcomes to the model alone.

\end{abstract}

\begin{center}\small Code, data and preregistrations: \url{https://github.com/nebula-1999/Interface-Induced-Trajectory-Censoring}\end{center}

\section{Introduction}
\label{sec:intro}

\begin{figure}[t]
\centering
\includegraphics[width=0.92\linewidth]{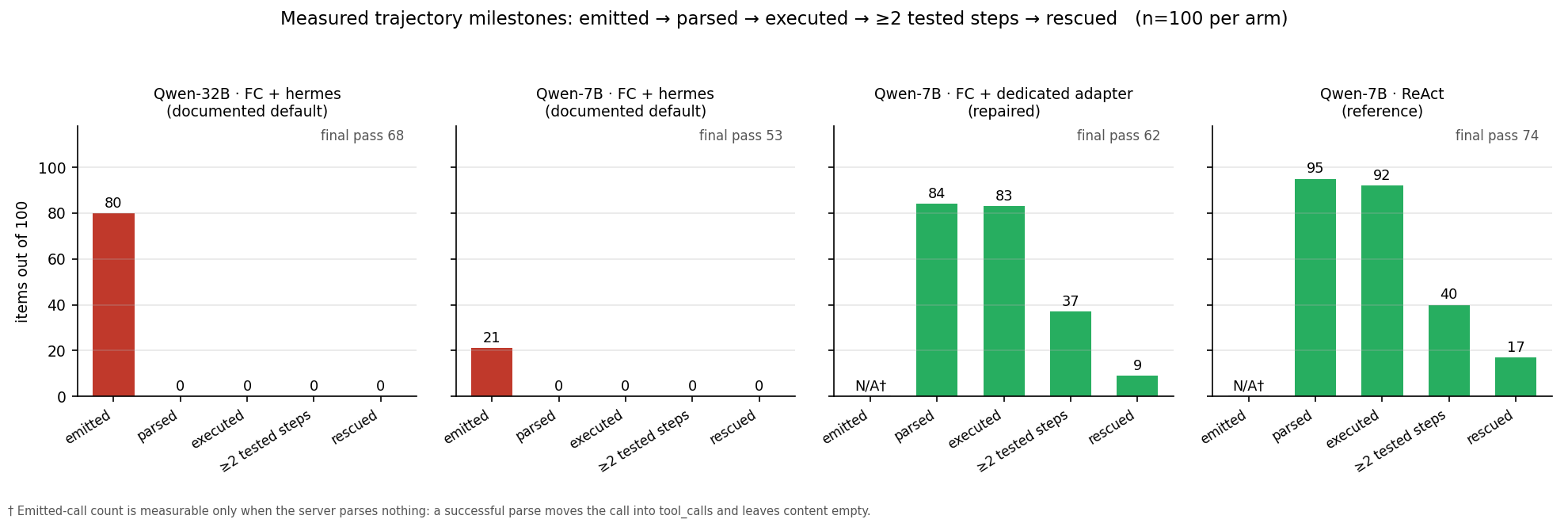}
\caption{Measured tool-use milestones ($n{=}100$ per arm). The fourth bar counts tasks reaching
at least two code-bearing tested steps (37 repaired; 40 ReAct), not returned observations.
N/A denotes cells that cannot be read from content
once the server has moved a successful call into \texttt{tool\_calls}.}
\label{fig:funnel}
\end{figure}

Agent benchmarks and training loops rarely consume raw model generations. They consume an action
produced by a chain: emission $\rightarrow$ serialization $\rightarrow$ parse $\rightarrow$
execution $\rightarrow$ observation $\rightarrow$ score or reward. A failure at several of these
boundaries has the same observable as a model declining to act: HTTP 200 and an empty
\texttt{tool\_calls} array. We call the resulting measurement failure \emph{interface-induced
trajectory censoring}. We distinguish masking of emitted actions from constructive suppression:
a decoding constraint can change what is emitted by disallowing invalid actions. The latter is
an intervention on generation, not evidence that a valid emitted action was censored. Figure~\ref{fig:funnel} shows the layers measured in our experiments.

Historical 1.5B runs motivated the investigation, but their ambiguous zero-execution result
does not identify parser masking (Appendix~\ref{app:extra}). We instead organize the evidence
around benchmark measurement, component interventions and the formal 7B control of training
experience, with explicit limits on what each observation identifies.

The individual Qwen incompatibility was reported previously in vLLM issue \#32926
\citep{vllm_issue_32926}. Our contribution is to quantify it as an evaluation-validity failure,
decompose its causes, and follow it into interactive evaluation and RL:
\begin{itemize}
\setlength{\itemsep}{1pt}
\item On BFCL v4, a template--parser $2\times2$ shows zero recovery from either one-component
replacement at the documented baseline and 196/200 parses when both are replaced; on
$\tau$-bench, the same joint repair restores 636 executions over 115 tasks.
\item A five-checkpoint ladder measures a growing absolute undercount under one mismatched
interface, while a preregistered matched-interface ladder keeps the silent count at 0--2 per 100.
\item A paired Llama intervention shows the opposite interface effect: relative to the
official-template baseline, \texttt{strict: true} suppresses 22 wrong-tool calls and raises final
pass from 44 to 61.
\item A preregistered 7B RL control changes only verl's parser registry. Repair restores 16,844
executed and observed calls, but neither the programmatic-feedback nor the post-hoc common native FC evaluation detects a success gain.
\end{itemize}

The resulting claim is narrower than ``the stack matters.'' A serving interface is part of both
the benchmark's measurement function and an agent's effective action space. Consequently,
tool-use results require layer-wise observability and a conformance check before model or protocol
attribution.

\section{Related Work}
\label{sec:related}

Evaluation outcomes vary with prompt format, benchmark choice, and other nuisance factors
\citep{sclar2024formatspread,mizrahi2024stateofwhatart,dehghani2021benchmarklottery}. Tool-use
audits similarly expose evaluator defects and ambiguous failure attribution
\citep{bhat2026benchmarkingbenchmarks,raj2026modelorharness,soni2026toolfailbench,albayaydh2026beyondleaderboard}.
Those studies generally vary inputs or inspect traces above the serving boundary. We study a
component inside the measurement apparatus: an independently configured serialization--parser
contract that can erase an action without an error. The closest analogue is a security-log study
where reparsing identical bytes changes measured accuracy from 0\% to 76\%
\citep{garware2026brokenruler}. We extend that structure to standard agent benchmarks, downstream
execution, and RL experience. Harness-Bench argues for reporting model--harness configurations
from cross-harness workflow measurements \citep{yao2026harnessbench}; we isolate one contract with
component-level interventions and follow it into training. Agent-Reactive Bugs finds silent
model--harness boundary failures in issue reports \citep{chen2026agentreactive}; our evidence is a
controlled intervention rather than a bug-report taxonomy. BFCL~\citep{bfcl2025}, $\tau$-bench~\citep{yao2025taubench}, and
other tool benchmarks~\citep{qin2024toolllm,li2023apibank} supply the evaluation context; verl
~\citep{sheng2025hybridflow}, GRPO~\citep{shao2024deepseekmath}, and LoRA~\citep{hu2022lora}
supply the training stack. Constrained generation can hurt reasoning
\citep{tam2024speakfreely}; our Llama control is a complementary case in which a constraint removes
role confusion and improves execution.

\section{Framework and Testable Signatures}
\label{sec:framework}

Let $Y$ be the assistant bytes produced from fixed weights, prompt, and decoding; $C(Y)$ an
independent indicator that those bytes contain a tool-intended action; $P(Y)$ the structured action
returned by the configured parser; and $E(P(Y))$ the execution and observation produced by the
environment. A benchmark normally records a function of $P(Y)$ or $E(P(Y))$, not $Y$. We call an
item \emph{masked} when $C(Y)=1$ but $P(Y)=\varnothing$. This definition avoids a metaphysical
``true capability'': the directly observed fact is that a behavior exists in the raw generation
and is absent from the structured trace. It also does not imply a parser bug. The parser may
correctly reject bytes outside its grammar; the invalid measurement arises when checkpoint,
template, and parser are treated as a compatible bundle without testing their contract.

The framework gives three diagnostic signatures. First, a parser-only intervention cannot change
bytes already generated, but can change execution, observations, later turns, and rescues. Our
evaluation-time Qwen repair changes parser, template, and few-shot together, so its identical
aggregate turn-1 count (53 versus 53) is a consistency check, not layer identification. We instead
localize the failure with fixed-output replay and event tracing. Llama's \texttt{strict:true}
intervention acts earlier by constraining generation and therefore can change turn-1 output.

Second, a size trend in $C(Y)-\mathbf{1}[P(Y)\neq\varnothing]$ requires both changing emissions
and a fixed rejection boundary. Across a comparable scale span, our preregistered Qwen3 ladder
under a matched envelope does not show the same growth. This supporting cross-family observation
does not isolate family from envelope and is not a strict scale counterfactual. Third, in RL the parser
changes the sampled experience distribution before a policy-gradient estimator sees it. Restoring
actions and observations establishes access to tool-mediated experience, not a measured
gradient contribution or eventual learning. The formal 7B control tests the two claims separately: event counts identify
experience access, while held-out checkpoints test whether that access changes behavior.

\section{Measurement Setup}
\label{sec:setup}
\label{sec:control}
\label{sec:gating}

Our controlled probe uses 100 KodCode problems~\citep{kodcode2025}, decontaminated against
EvalPlus by normalized text/code n-gram containment, paired across arms, with one
\texttt{run\_tests(code)} tool. Function calling uses
\texttt{tool\_choice:auto}, temperature 0, and 2,048 output tokens. We separately run BFCL v4 at
pinned Gorilla commit \texttt{6ea5797} and all 115 $\tau$-bench \texttt{retail} tasks. The latter
uses a locally served Llama-3.1-8B user simulator rather than the original harness's
\texttt{gpt-4o}; its absolute score is therefore not leaderboard-comparable, and we claim only
paired arm differences.

Serving uses vLLM 0.27.1~\citep{kwon2023vllm}. Server-parsed calls are insufficient for the
question under study. For the KodCode probe, a shared classifier marks a
generation \emph{tight} using a regex for a call-shaped payload naming \texttt{run\_tests},
a quoted \texttt{arguments.code} field and Python-like lexical markers. This heuristic rejects
some schema echoes and placeholders but does not validate JSON syntax or Python executability. Against a third-party-adjudicated gold standard its $\kappa$ is 0.871, and tight-stratum
precision is 36/40=0.900. Two annotators achieve $\kappa=0.967$ on the 62 items presented with
identical bytes. Stored evaluation outputs have a 4,000-character cap, which creates false
negatives, while 0.900 precision establishes false positives; the net bias is therefore not
identified. In the stratified validation sample, cap hits by scale are 0/6, 0/13, 0/15, 0/24,
and 10/40 from 1.5B through 32B, so truncation uncertainty is concentrated at the headline rung. Across all 100 stored first outputs per scale, cap hits are
0, 0, 1, 0 and 18; a cap hit does not reveal the omitted continuation.
Formal RL rollouts are retained in full with hashes. Full validation appears in
Appendix~\ref{app:humanval}.

Before interpreting a zero we verify that \texttt{tool\_choice:required} produces a correctly
named, parseable call on the same live server where this control is available; required-mode
behavior is not assumed identical to auto-mode parsing. This positive control is distinct from the zero-error admissibility gate; zero auto-mode parses are not an exclusion criterion. An arm enters the clean-arm pass-rate analysis only with 100
items, exit code 0, zero request errors, and matching provenance; this gate is enforced in code.
End-to-end analyses are reported separately using full task denominators, including request failures.
Appendix~\ref{app:errata} records an earlier invalid $p=0.648$ that this rule forced us to withdraw.

For paired binary outcomes we use exact two-sided McNemar tests and report both discordant counts;
for the formal RL endpoint we report a conservative finite-sample 95\% paired-risk interval
formed from exact binomial bounds on the two discordance probabilities. It reflects item-pair
uncertainty, not seed or inference-retest variation. The historical retrospective family has 27 tests
($0.05/27\approx0.00185$); nine later native FC contrasts are separately exploratory
(Appendix~\ref{app:statistics}). Neither policy was preregistered. We
distinguish a mechanism count (parse, execution, observation) from a task outcome throughout: a
large deterministic movement in the former does not license a performance claim when the paired
outcome test is null. BFCL scores come from its official scorer; $\tau$-bench outcomes come from
final environment state; code-task outcomes execute the generated program against held-out tests.
Training events exclude evaluation requests. Historical 1.5B curves exclude two verified
EvalPlus defects (\texttt{HumanEval/32} and \texttt{Mbpp/599}) and use $n=540$. The preregistered
7B programmatic-feedback evaluation retains all 542 items; both flagged items fail in both arms.
The later native FC evaluation scores them differently; Appendix~\ref{app:nativefc}
reports its full denominator and a 540-item sensitivity analysis.

\section{Results}
\label{sec:results}

\subsection{External validity: BFCL and $\tau$-bench}
\label{sec:bfcl}
\label{sec:tau}

Using Qwen2.5-Coder-7B-Instruct for both external benchmarks, and holding weights, cases,
decoding, seeds, executor, and scorer fixed on BFCL v4, the
documented Hermes configuration yields official scores 0.00 on both \texttt{simple\_python} and
\texttt{multi\_turn\_base}; the dedicated Qwen2.5-Coder adapter yields 0.96 and 0.19. Because the
adapter changes both template and parser, we ran the full $2\times2$ (Table~\ref{tab:bfcl2x2}).
Changing either component alone from the documented baseline recovers none (0/200); changing both
recovers 196/200. Thus the two one-factor simple effects at that baseline are zero, while the joint
replacement shows strong complementarity. This statement concerns parse counts; official BFCL
scores were computed for the two end-to-end corner configurations, not all four cells. Each
component behaves according to its own contract, but the contracts disagree. The documented arm has zero HTTP errors and
166/200 responses satisfying the BFCL JSON-call criterion despite zero parses. On \texttt{multi\_turn\_base}, repair moves execution from
0 to 98/100 and success from 0 to 19/100. Complete scorer and funnel tables are in
Appendix~\ref{app:extra}. ``Documented'' refers to applying Qwen2.5 family-level guidance,
not a Coder-specific guarantee (Appendix~\ref{app:config}). A new, answer-blind offline JSON
extractor applied to the original 100 \texttt{simple\_python} responses recovers 97 structurally
schema-compliant calls; 92 pass the pinned official AST checker without regenerating text
(Appendix~\ref{app:fixedreplay}). This is distinct from the deployed dedicated parser.

\begin{table}[H]
\caption{BFCL first-request parses, template $\times$ parser ($n=200$).}
\label{tab:bfcl2x2}
\centering
\small
\begin{tabular}{lcc}
\toprule
 & documented template & dedicated template \\
\midrule
Hermes parser & 0 & 0 \\
Dedicated parser & 0 & \textbf{196} \\
\bottomrule
\end{tabular}
\end{table}

On paired $\tau$-bench tasks, the same adapter swap moves parsed and executed calls from
0 to 636 and tasks reaching any execution from 0/115 to 103/115. All 1,389 documented-arm
requests carry tools; 789 assistant turns contain call-shaped payloads that are not parsed.
Replay assigns all 789 turns to the missing-envelope category; none reaches the
payload-validation stage, and none is accepted by the configured extractor. Outcomes move from 7 to 10 solved, with
three discordant pairs in the repaired direction ($p=0.25$); we make no outcome claim. This result
extends the envelope-layer mechanism to a non-code, stateful environment, but not the scale or RL
claims, which remain code-only (Table~\ref{tab:external_main}).

\begin{table}[H]
\caption{External benchmarks under documented and repaired adapters. BFCL rows are official
scores; $\tau$-bench rows are paired counts over 115 tasks.}
\label{tab:external_main}
\centering
\small
\begin{tabular}{lrr}
\toprule
measurement & documented & repaired \\
\midrule
BFCL \texttt{simple\_python} score & 0.00 & \textbf{0.96} \\
BFCL \texttt{multi\_turn\_base} score & 0.00 & \textbf{0.19} \\
BFCL multi-turn executed/succeeded & 0/0 & \textbf{98/19} \\
\midrule
$\tau$-bench parsed/executed calls & 0/0 & \textbf{636/636} \\
$\tau$-bench tasks reaching a tool & 0 & \textbf{103} \\
$\tau$-bench tasks solved & 7 & 10 \\
\bottomrule
\end{tabular}
\end{table}

\subsection{Scale under mismatched and matched interfaces}
\label{sec:scale}

Under the documented Hermes interface, the Qwen2.5-Coder server reports zero calls at all five
sizes while tight-classified emissions rise monotonically from 0 to 80/100
(Table~\ref{tab:ladders}). A sensitivity estimate is approximately 72 if tight-stratum
precision transfers across sizes; this does not restore false negatives or estimate all true calls. A random 300-item
replication gives 0, 4, 30, 40, and 81 per 100: the 7B rung moves from 21 to 30, but direction and
monotonicity remain. This is a within-family observation under one mismatch, not a scaling law;
the server column is constant, so the slope comes entirely from increasingly frequent or regular
emissions. The 4,000-character cap creates false negatives, while classifier precision below one
creates false positives; the raw count's net bias is not identified.

The preregistered supporting ladder uses Qwen3 checkpoints up to 8B whose required envelope matches Hermes.
The prediction ``parsed $>0$ at every size'' was committed before inference. It holds, while the
silent count stays 1, 1, 2, and 0 through 8B; the separate mismatched ladder reaches 80 at
32B. This cross-family observation is
consistent with contract mismatch, rather than size alone, driving the undercount; it does not
isolate the envelope because family and envelope change together. Parsed counts are not monotone, and we disabled thinking because the
default exhausted the token budget before a call; both facts were recorded as protocol deviations.
An attempted Granite rescue produced no calls under either Hermes or its native parser and is
uninformative, not confirmatory.

\begin{table}[H]
\caption{Scale ladders. Tight counts concern unparsed outputs only (including Qwen3); they are not total emitted-call counts. ``Silent'' denotes these classifier-positive, unparsed outputs.}
\label{tab:ladders}
\label{tab:scale}
\centering
\small
\begin{tabular}{llrr}
\toprule
family/interface & size & parsed & tight, unparsed / 100 \\
\midrule
Qwen2.5-Coder/Hermes & 1.5B & 0 & 0 \\
 & 3B & 0 & 4 \\
 & 7B & 0 & 21 \\
 & 14B & 0 & 36 \\
 & 32B & 0 & 80 \\
\midrule
Qwen3/matched & 0.6B & 86 & 1 \\
 & 1.7B & 44 & 1 \\
 & 4B & 70 & 2 \\
 & 8B & 97 & 0 \\
\bottomrule
\end{tabular}
\end{table}

A within-lineage Qwen2.5-Instruct control further localizes the effect: under the same template,
parser, items, and decoding, parsed rates are 1, 11, 63, 77, and 88 rather than Coder's flat zero.
After rerunning with a working executor, multi-turn rescues rise 0, 3, 6, 14, and 23 with that
parsed column. The contrast is consistent with checkpoint-specific post-training differences,
not parameter count or the shared family/template alone; it does not identify tool-format training
as the unique cause. Across 4,254 de-duplicated first turns from all archived arms, replay finds
zero cases that vLLM's configured parser should have accepted but did not: this is contract
mismatch, not a parser implementation bug (Table~\ref{tab:instruct_main}).

\begin{table}[H]
\caption{Within-lineage Qwen2.5-Instruct control under the same interface and items. Rescues are
final passes not passing on turn 1.}
\label{tab:instruct_main}
\centering
\small
\begin{tabular}{lrrrr}
\toprule
size & parsed & turn-1 & final & rescues \\
\midrule
1.5B & 1 & 13 & 13 & 0 \\
3B & 11 & 14 & 17 & 3 \\
7B & 63 & 34 & 40 & 6 \\
14B & 77 & 38 & 52 & 14 \\
32B & 88 & 42 & 65 & 23 \\
\bottomrule
\end{tabular}
\end{table}

\subsection{Llama: a constraint suppresses the wrong action}
\label{sec:llama}
\label{sec:taxonomy}
\label{sec:families}

Llama-3.1-8B calls the function requested by the programming problem rather than
\texttt{run\_tests} on 23/100 items. A richer schema leaves 22 wrong calls (discordances 6/5,
$p=1.0$), a thought scaffold worsens them to 59 ($p=2.84\times10^{-10}$), and the official
template leaves 22 (1/0, $p=1.0$). The previously quoted $p=0.125$ belongs to the final-pass
comparison, not wrong-tool labels. Adding \texttt{strict:true} to the official-template arm
reduces wrong calls from 22 to 0 (22/0, $p=4.77\times10^{-7}$), raises valid execution from 73
to 97, and final pass from 44 to 61, a 17-point single-variable change. The broader terse-FC to
official-plus-strict comparison used in the ReAct decomposition is 49 to 61, hence 12 points for
the interface bundle ($p=0.0075$), followed by 19 for protocol ($p=0.0019$). These are descriptive score differences;
neither contrast survives the retrospective family-wise correction. The official-template
strict final-pass comparison does survive that correction. Thus the model exhibits
role confusion under unconstrained function calling, while the interface determines whether that
confusion reaches the environment. Appendix~\ref{app:errata} corrects the corresponding initiation
rate from 97 to 74. DeepSeek and Mistral only motivate additional template/token failure layers:
the former never injects tools, and all four Mistral FC variants contain request errors, so neither
supports a comparable performance claim.
Table~\ref{tab:llama_main} reports the paired wrong-tool tests and their comparison baselines.

\begin{table}[H]
\caption{Llama-3.1-8B paired controls. The first three interventions do not fix role confusion;
\texttt{strict:true} does.}
\label{tab:llama_main}
\centering
\small
\begin{tabular}{lrr}
\toprule
arm & wrong-tool calls & exact paired $p$ (baseline) \\
\midrule
terse FC & 23 & --- \\
rich schema & 22 & 1.000 (terse) \\
thought scaffold & 59 & $2.84\times10^{-10}$ (terse) \\
official template & 22 & 1.000 (terse) \\
official template + strict & \textbf{0} & \textbf{$4.77\times10^{-7}$ (official)} \\
\bottomrule
\end{tabular}
\end{table}

\subsection{Repair changes evaluation trajectories}
\label{sec:repair}
\label{sec:maintable}

On Qwen2.5-Coder-7B, changing only the evaluation adapter moves parses 0$\rightarrow$84, items
reaching a second code-bearing tested step 0$\rightarrow$37, rescues 0$\rightarrow$9, and final pass
53$\rightarrow$62. Aggregate turn-1 pass is 53 in both arms, while Llama's schema intervention
moves turn-1 pass from 33 to 46 on the official template. Because the Qwen adapter also changes template and few-shot, this
aggregate pattern is descriptive rather than a failure-layer locator; fixed-output replay and
event tracing provide that localization. The Qwen final gain is not significant at
$n=100$ (16 gains / 7 losses, $p=0.093$) or after the random $n=300$ replication and a Bonferroni threshold of
approximately 0.00185 ($160\rightarrow178$, nominal $p=0.0385$). Repair restores the mechanism;
we do not claim an evaluation outcome effect (Table~\ref{tab:repair_main}).

\begin{table}[H]
\caption{Qwen2.5-Coder-7B evaluation repair; only the serving adapter changes.}
\label{tab:repair_main}
\centering
\small
\begin{tabular}{lrrrr}
\toprule
arm & parsed & $\geq2$ tested steps & turn-1/final & rescues \\
\midrule
FC + Hermes & 0 & 0 & 53/53 & 0 \\
FC + dedicated adapter & \textbf{84} & \textbf{37} & 53/62 & \textbf{9} \\
ReAct reference & 95 & 40 & 57/74 & 17 \\
\bottomrule
\end{tabular}
\end{table}

\subsection{Parser repair restores RL experience; no held-out improvement detected}
\label{sec:rl}
\label{sec:norepair}
\label{sec:sufficiency}

The RL experiment is a preregistered single-variable 150-step 7B GRPO comparison. Both arms
are evaluated by the same parser-independent protocol, which automatically executes generated
code and supplies test feedback; this is not an evaluation of voluntary tool initiation. Both arms use the same
LoRA rank 32, data, mandatory prompt, seed, decoding, verifier, evaluation, and training budget;
the sole intervention is verl's registered parser. Full emissions, parser verdicts, executions,
observations, and hashes are retained. Table~\ref{tab:p3formal} excludes held-out evaluation from
training event counts and uses the same paired 542-item evaluation set at every checkpoint. Both
step-0 evaluations use the same base checkpoint and evaluator; their equal aggregate counts are
an observation, not evidence from which we infer the evaluation design.

\begin{table}[H]
\caption{Formal 7B RL control. Tight counts generation events; accepted counts parsed calls; exec. counts tool invocations, all observed. Final/rescues count evaluation items.}
\label{tab:p3formal}
\centering
\small
\begin{tabular}{lrrrr}
\toprule
arm & tight & accepted & exec. & final/rescues, step 0$\to$150 \\
\midrule
broken FC & 8,689 & 10 & 10 & 430$\to$430 / 12$\to$12 \\
repaired FC & 17,608 & 16,912 & 16,844 & 430$\to$427 / 12$\to$12 \\
\bottomrule
\end{tabular}
\end{table}

The 16,912 accepted calls occur in 16,844 generation events: 40 generations contain multiple
calls and contribute 68 calls beyond the first. The configured one-call cap selects the first
parser-returned call and leaves the remaining calls undispatched, not merged. All 16,844
dispatches join to executions and observations. Extraction-to-dispatch counts match per trajectory,
not through per-call IDs; Appendix~\ref{app:p3curve} details this audit boundary.

The broken channel is nearly, not literally, closed: 8,689 tight-classified generation events and
only 10 executions are recorded. Repair changes sampled tool experience by more than three orders of magnitude and returns
an observation for every execution. Under the common programmatic-feedback evaluation, no held-out improvement is detected: repaired FC ends at
427/542 versus 430/542 for broken FC (paired difference $-0.55$ percentage points; conservative
paired 95\% CI $[-1.79,0.72]$; all three discordances favor broken; exact McNemar $p=0.25$), and
rescues are 12 at both initialization and the endpoint; intermediate checkpoints vary. Relative to its own initialization, repaired FC has one gain and four losses
($p=0.375$). We do not interpret these non-significant contrasts as harm. At this scale, budget,
and single seed, parser repair restores tool-mediated experience, but we detect no improvement in
held-out success or rescue counts. Held-out evaluation is identical across arms and deliberately
bypasses both training parsers: the current LoRA is queried through verl's native rollout RPC at
temperature 0 for up to four 1,024-token programmatic-feedback turns, generated code is executed
on the same 542 EvalPlus items, and test output is appended as a user message. Training uses
structured tool-role feedback; this format shift may limit transfer, alongside adaptation
capacity, training budget and credit assignment. This protocol does not measure native FC initiation. This explains why
the broken-training arm can still show 12 evaluation rescues. Each evaluation request attests the
loaded global weight step.

A post-hoc common native FC evaluation of all three weights on the same 542 items
also detects no repaired-versus-broken success gain: both pass 248/542 versus 239/542
for base (16 discordances each way; difference 0 points, conservative 95\% CI
$[-3.47,3.47]$; $p=1$). Accepted-call initiation occurs on 46/50/47 tasks for base/broken/repaired;
tool-test rescues are 2/1/1. Only accepted calls receive tool-role feedback; terminal scoring
is separate. This is not an equivalence result or an additional training seed
(Appendix~\ref{app:nativefc}).

The repaired job was interrupted after step 116 and
resumed from a validated step-90 checkpoint; model, optimizer, scheduler, RNG, dataloader, source,
configuration, and weight-step hashes were checked. The repeated step-90 evaluation differs by
two final-pass items, reported as test--retest variation. Formal event totals use the final model's
lineage---original steps 1--90 plus resumed steps 91--150---and exclude the abandoned 26-update
step-91--116 branch. Resume evidence and the available checkpoint endpoints are in
Appendix~\ref{app:p3curve}.

Two details prevent common but incorrect readings of this comparison. Both arms use LoRA because
7B full-parameter GRPO does not fit on one 80\,GB GPU; this preserves the arm contrast but prevents
strict comparison with the historical 1.5B full-parameter curves. Event totals also differ because
repair changes trajectories: once calls execute, conversations contain more turns and hence more
opportunities to emit another call. We therefore use rates within each arm only to describe its
funnel, and use the single registered-parser intervention for causal attribution, not equality of
raw denominators. The old ReAct arm with 23,676 executions and flat rescues is directionally
consistent, but remains supporting evidence because it changes protocol as well as interface.

\section{Implications and Limitations}
\label{sec:discussion}
\label{sec:limitations}
\label{sec:future}

\paragraph{Measure the funnel.} Reports should include emitted-call rate under a stated criterion,
parser acceptance, dispatch/execution, observation return, request errors, adapter/template/parser
versions and hashes, and task outcome. We provide a 98-line preflight that sends a canonical
tools-bearing request, validates the returned name and arguments, and repeats under
\texttt{tool\_choice:required} as a positive control. This screening check flags configurations
that fail to return an expected tool call; it does not by itself identify the underlying
failure mechanism. Generation constraints are a distinct potential remedy, altering emissions rather than
reparsing existing bytes; they do not by themselves ensure task success. This should become a machine-readable conformance report rather than a model-only call rate.

At minimum, such a report should persist model and tokenizer revisions, template and parser
hashes, decoding arguments, request errors, the emitted-intent criterion, acceptance, dispatch,
execution and observation counts, and the outcome delta. Most fields already exist somewhere in a
normal trace. The standard changes the unit of reporting from ``model call rate'' to a versioned
model--interface measurement card, making silent zeros auditable and adapter changes comparable.

\paragraph{Interpretation limits.} The scale and RL results use one code task and one tool; only
the envelope mechanism is replicated on a non-code benchmark. The scale trend is one family under
one mismatch, with single-sample headline arms and a 32B ceiling. The RL result is one 7B LoRA seed
and budget, and near-zero samples do not establish zero policy probability. The historical 27-test family
has a Bonferroni threshold of 0.00185; only the strongest interventions survive it, and we
rest no claim on the nominal or null contrasts. A historical verifier used an unanchored pytest
regex and omitted skipped/expected-failure outcomes; the formal arms use the same repaired verifier
and retain full rollouts, so that flaw cannot explain their contrast. Serving/training results depend on vLLM 0.27.1
and verl 0.9.0; the later native FC evaluation uses Transformers 5.10.4 and PEFT 0.20.0.
Appendix~\ref{app:limitations} states all 17
limitations, including the non-random sample, post-treatment conditioning, asymmetric family
taxonomy, human-validation scope, provenance corrections, and the observations that would change
our conclusions.

Several of those limits bear directly on the headline. The original 100 items are
\texttt{clean[:100]}, not a random draw; the random 300-item replication preserves the ordering
but moves the 7B rung from 21 to 30 per 100. Conditioning protocol comparisons on items parsed by
both arms is post-treatment selection, so the Qwen residual of +8.4 points
($95\%$ CI $[-0.5,17.4]$, $p=0.118$) is unresolved rather than evidence of equivalence. The
taxonomy is intentionally asymmetric: Qwen and Llama have controlled repairs, while Mistral has
no admissible 100-item FC run and DeepSeek only a template diagnosis. Complete RL
attempt-versus-discard attribution exists only for the new 7B runs. The historical 1.5B data
remain over-determined and are not used for the causal training claim.

\paragraph{What would change the conclusions.} A replicated or longer repaired-FC run with a
consistent rescue gain would narrow the scope of the present 150-step, single-seed null result. A second family
whose undercount does not grow with size would confine the scale observation to Qwen-plus-Hermes;
a comparable rise would strengthen the scaling interpretation. A non-code size ladder would
separate the envelope mechanism already seen on $\tau$-bench from the code-shaped payloads used in
the current ladder. A second serving stack would identify which findings are vLLM-specific. These
tests and decision rules are stated in full in Appendix~\ref{app:limitations}, rather than
presented as evidence already obtained.

\section{Conclusion}

A tool-call rate measures a model--serialization--parser--execution composition. On BFCL, neither
single-component replacement restores parsing from the baseline, whereas their joint replacement moves the official score from 0.00
to 0.96; in RL, changing only the parser moves observed executions from 10 to 16,844 without a
detected gain under either common evaluation protocol. The practical consequence is simple: verify the contract and report each
funnel layer before attributing a terminal zero, a protocol gap, or a training curve to the model.

\subsection*{AI use statement}
Generative AI systems assisted with experiment design feedback, implementation and debugging,
result interpretation, literature discovery, and manuscript drafting and editing. The authors
inspected the underlying trajectories and logs, reran validation checks, verified every reported
number against persisted artifacts, and manually reviewed the final claims and citations. No
AI-generated output is treated as experimental evidence. The authors take responsibility for all
content, including code, claims, and text produced with AI assistance.

\subsection*{Reproducibility statement}
Code, analysis scripts, preregistration documents, raw BFCL and $\tau$-bench records, the formal RL audit summary and the native function-calling outcome records are in the public repository (\url{https://github.com/nebula-1999/Interface-Induced-Trajectory-Censoring}). A supplementary package, available from the authors on request, contains executable statistical code, reduced outcome inputs,
component versions and configuration summaries, file hashes, retained preregistration documents,
full statistical outputs, reduced per-item RL evaluations, and compact event records including
excluded resume segments. Raw BFCL and $\tau$-bench records plus a deterministic RL text sample
support content inspection and single-turn BFCL replay. The locally
retained full rollout archive is not part of that package; that disclosure boundary
is explicit in its README. Appendix~\ref{app:repro}
provides a minimal reproduction path; Appendices~\ref{app:config}--\ref{app:extra} provide configurations,
arm inventories, validation records, and complete result tables. Appendix~\ref{app:errata} records every
known provenance or instrumentation error and its effect on the reported conclusions.

\bibliography{refs}
\bibliographystyle{vendor/iclr2027_conference}

\appendix
\section{Data Errata}
\label{app:errata}
\emph{Every number in this paper traces to a trajectory file under
\texttt{runs/final/}. This appendix lists the files whose recorded provenance is wrong, the
arms that are inadmissible for pass-rate comparison, and what was verified correct. It is
reproduced from \texttt{runs/final/ERRATA.md} in the released repository.}

\subsection{Seven ReAct arms record the wrong \texttt{max\_tokens}}

The following files record \texttt{max\_tokens=2048} while the generation limit in force
was \textbf{1024}. Cause: \texttt{gen()} in the probe carried a default of
\texttt{max\_\allowbreak tokens=1024} while \texttt{gen\_fc()} used the global \texttt{MAX\_TOKENS=2048},
and the provenance written to disk recorded \texttt{MAX\_TOKENS} for both. Fixed
2026-08-31.

{\footnotesize
\begin{verbatim}
traj_v3_Llama8B_react.jsonl        recorded 2048 / actual 1024
traj_v3_Qwen7B_react.jsonl         recorded 2048 / actual 1024
traj_v3_Mistral7B_react.jsonl      recorded 2048 / actual 1024
traj_v11_DS1.3b_react.jsonl        recorded 2048 / actual 1024
traj_v11_DS6.7b_react.jsonl        recorded 2048 / actual 1024
traj_v11_Llama32_1B_react.jsonl    recorded 2048 / actual 1024
traj_v11_Llama32_3B_react.jsonl    recorded 2048 / actual 1024
\end{verbatim}
}

\textbf{Direction of the bias.} The contemporaneous FC arms were genuinely at 2048, so in
every affected ReAct-vs-FC comparison \textbf{the function-calling side had the larger
generation budget}. A larger budget need not improve performance, so this asymmetry alone
does not establish the direction of bias. Re-running the three
main ReAct arms at a true 2048 changed final pass by at most one item
($80\to80$, $74\to74$, $32\to33$), which is why the correction is small in practice.

\textbf{Replacement data.} The four v11 arms are superseded by
\texttt{traj\_v13\_*\_react.jsonl}; the three v3 arms by \texttt{traj\_v14\_*\_react.jsonl}.
All tables in this paper cite the v13/v14 versions; the v3/v11 ReAct arms are retained only
as a record.

\subsection{Llama-3.1-8B initiation rate required manual correction}

\texttt{traj\_v3\_Llama8B\_fc.jsonl} records an initiation rate of 97/100, but 23 of those
calls target \textbf{the task function rather than \texttt{run\_tests}}. The probe did not
verify tool names when that batch was collected, so \texttt{has\_action} did not distinguish
them. \textbf{The corrected \texttt{run\_tests} initiation rate is 74/100.}

A re-run restricted to exactly those 23 items under matched configuration
(\texttt{traj\_\allowbreak v8\_\allowbreak Llama8B\_\allowbreak recheck.jsonl}, seed fixed) reproduced 23/23 as wrong-tool
calls, with names such as \texttt{can\_\allowbreak form\_\allowbreak word}, \texttt{check\_\allowbreak password\_\allowbreak strength} and
\texttt{floyd\_warshall}.

Of the same batch, \texttt{traj\_v3\_Qwen7B\_fc\_nosuffix.jsonl} has an initiation rate of
0 and needs no correction; \texttt{traj\_v3\_Mistral7B\_fc.jsonl} has 2, and those two were
not individually re-checked for tool name.

\subsection{Arms inadmissible for pass-rate comparison}

{\footnotesize
\begin{verbatim}
traj_v3_Mistral7B_fc.jsonl                  n_err = 2   (rc=2)
traj_v5_Mistral7B_fc_official.jsonl         n_err = 42  (rc=2)
traj_v9_Mistral7B_fc_strict.jsonl           n_err = 3   (rc=2)
traj_v9_Mistral7B_fc_official_strict.jsonl  n_err = 39  (rc=2)
traj_v13_Llama8B_fcstrict_t06_s3.jsonl      n_err = 1   (rc=2)
\end{verbatim}
}

For these arms the \textbf{error-rate census remains valid} --- that census is itself a
result --- but these arms are excluded from clean-arm pass-rate comparisons, and no
clean-arm $p$-value is computed for them. Separately reported end-to-end analyses retain
the full task denominator, including request failures. An earlier draft computed $p=0.648$ on one such arm; that number is
retracted.

\subsection{Provenance schema drift}

Fields were added over the course of the work, giving four schema generations: early arms
lack \texttt{seed}, \texttt{temperature} or \texttt{fc\_schema}. The fields load-bearing for
every claim (\texttt{model}, \texttt{protocol}, \texttt{adapter}, \texttt{max\_tokens},
\texttt{clean\_index}) are present in all generations. A \texttt{script\_sha256} field was
never successfully added --- the patch was interrupted twice --- so script identity across a
run is instead guaranteed externally, by hashing the probe before launch
(\texttt{v13\_pinned\_hashes.txt}) and re-checking after.

\subsection{Verified correct}

\begin{itemize}
\tightlist
\item \textbf{All 50 full-length arms use the identical 100-item set}
      (\texttt{clean[:100]}; unique item-set count $=1$), so every cross-arm comparison in
      this paper is exactly paired.
\item Every FC arm ran at a true \texttt{max\_tokens} of 2048.
\item Sampling in the variance experiment is genuine: at \texttt{temperature}~$=0.6$, two
      seeds differ in first-turn code on 95/100 items and flip outcomes on 16/100.
\end{itemize}

\subsection{Third-party component versions}

The Qwen2.5-Coder \texttt{<tools>} parser is taken from
\path{hanXen/vllm-qwen2.5-coder-tool-parser} (Apache~2.0), downloaded 2026-08-31, at the
then-current \texttt{main}:

{\footnotesize
\begin{verbatim}
commit 1b921501f30cbfe347dccb1db7de3c82a1d55131
  (1b92150, 2026-04-29)
  "fix: buffer partial <tools> prefix in streaming
   to prevent tag leak"

SHA256 (first 20):
  c16bb1f88936a2d96c7c  qwen2_5_coder_tool_parser.py
                      (unmodified)
  736bd175adbf90942c1d  tool_chat_template_qwen2_5_coder.jinja
                      (MODIFIED)
  a95b2a9b91e65b3d452f  tool_chat_template_qwen2_5_coder.jinja.orig
                      (original)
\end{verbatim}
}

The template was modified because its few-shot examples invoke \texttt{get\_weather}, a
tool absent from our schema. Left unchanged it would induce calls to a non-existent tool,
simultaneously inflating the initiation count and contaminating the empty-argument
statistic. The original is retained alongside.

\section{Reproducibility}
\label{app:repro}

The formal 7B configuration and shared held-out protocol are in Appendix~\ref{app:config};
the checkpoint curve and accepted resume lineage are in Appendix~\ref{app:p3curve}.
Appendix~\ref{app:statistics} defines the interval and retrospective test family.
The supplementary package contains reduced Llama outcomes, 11{,}952 reduced 7B evaluation
records, compact events including excluded segments, lineage metadata and verification
scripts. It also includes a small raw-content mechanism archive: 800 first responses from
the BFCL factorial, 100 documented and 100 repaired single-turn BFCL records with request
schemas and official-scoring inputs, all 230 $\tau$-bench task trajectories, and ten
deterministically selected step-1 RL trajectories per arm. The full training rollout archive,
original evaluation programs, training logs and heavy checkpoints are not included.

\begin{center}
\small
\begin{tabular}{@{}p{0.70\linewidth}p{0.23\linewidth}@{}}
\toprule
Verification target & Package support \\
\midrule
Llama paired tests from reduced item records & Yes \\
Other historical tests from stated discordances & Yes, counts only \\
7B checkpoint outcomes and endpoint pairing & Yes \\
7B recorded funnel and accepted resume segments & Yes \\
Inspect/replay supplied BFCL raw outputs & Yes \\
Inspect $\tau$-bench and sampled RL raw contents & Yes \\
Reclassify every training emission & No \\
Verify checkpoint weights or raw text hashes & No \\
\bottomrule
\end{tabular}
\end{center}

From the extracted package root, run:
\begin{verbatim}
python3 analysis/verify_supplement.py
python3 analysis/submission_statistics.py --output /tmp/stats.json
python3 analysis/bfcl_fixed_replay.py \
  --bundle mechanism_evidence/bfcl
\end{verbatim}
The event verifier uses the Python standard library and reconstructs inclusion from
segment/step ranges. Dispatch, execution and observation can be uniquely joined.
Extraction lacks per-turn request identity, so accepted-to-dispatch matching is only
at trajectory-count granularity. Sampled RL events preserve raw emissions and recorded
observations, but omit training task inputs and executable tests: they support text inspection,
not independent environment re-execution. BFCL request fields come from its inference log,
not a byte-for-byte HTTP request capture. Hashes do not substitute for file contents.

Full-archive-only procedures include \texttt{analysis/p3\_final\_evidence.py},
\texttt{analysis/intent.py}, \texttt{runs/final/final\_table.py}, checkpoint audits
and historical figure regeneration. They require data not supplied in the supplementary
package and are not its default verification workflow. Historical trajectories and
their configuration index reside under \texttt{runs/final/} in the full research
archive; formal 7B records are retained separately. Only explicitly listed reductions and
raw-content subsets are provided in the package.

Model snapshots and serving configurations must be pinned because template inheritance
can change with revision. The third-party parser is hanXen (\texttt{1b92150}, Apache 2.0);
paired few-shot examples were rewritten from \texttt{get\_weather} to \texttt{run\_tests}.
Original and adapted hashes are retained in the full archive.

The post-hoc native FC evaluation adds \texttt{native\_fc/outcomes.json}
(1,626 reduced item records), \texttt{native\_fc/statistics.json}, and
\texttt{analysis/native\_fc\_results.py}. From the package root run
\texttt{python3 analysis/native\_fc\_results.py --outcomes native\_fc/outcomes.json}
\texttt{--out /tmp/native-fc-recheck}. This reproduces paired contrasts and the
timeout inventory; it does not re-execute generated programs. Full raw conversations
and GPU loading receipts are retained separately, not promised as part of this reduction.

\section{Prompts and tool schemas: English renderings}
\label{app:prompts}

The experiments used Chinese prompts. To keep this manuscript entirely in English, this
appendix gives faithful English renderings; the byte-exact Chinese originals are reproduced
verbatim in the extended manuscript in the public repository
(\texttt{paper/sections/C\_prompts.tex}). The renderings below were never sent to a model and
are provided only for readers.

\subsection{ReAct prompts}

\begin{Verbatim}[fontsize=\small,breaklines=true,breakanywhere=true]
REACT_OPTIONAL
--------------
You are a Python coding assistant. You may work in the following format:

Thought: your reasoning
Action: run_tests
Action Input: ```python
<complete code>
```
Observation: (filled in by the system)

Thought: I have confirmed that the code is correct
Final Answer: ```python
<final code>
```

You may submit code for verification with Action, or answer directly.
\end{Verbatim}

The mandatory variant replaces the final sentence with: ``You must first submit code for
verification with Action, then give a Final Answer,'' and marks the loop as repeatable. The main
evaluation uses the optional prompt for both protocols. Historical FC--ReAct training and both
formal P3 arms use mandatory prompts of matched strength.

\subsection{Function-calling prompts}

\begin{Verbatim}[fontsize=\small,breaklines=true,breakanywhere=true]
FC_OPTIONAL
-----------
You are a Python coding assistant. Implement the function described by the user.
You may call run_tests to submit your code to the test runner. It returns
pass/fail status and errors, which you may use to revise and resubmit.

FC_MANDATORY
------------
You are a Python coding assistant. Implement the function described by the user.
You must first call run_tests to verify your code; do not answer directly.
The tool returns pass/fail status and errors. If any test fails, revise the
code and call run_tests again.
\end{Verbatim}

\subsection{Terse, rich, and strict schemas}

The terse schema is the documentation-level baseline. Its English rendering is:

\begin{Verbatim}[fontsize=\small,breaklines=true,breakanywhere=true]
{"type": "function", "function": {
  "name": "run_tests",
  "description": "Submit the Python code you wrote to the test runner; it
                  returns pass/fail status and errors.",
  "parameters": {"type": "object",
                 "properties": {"code": {"type": "string"}},
                 "required": ["code"]}}}
\end{Verbatim}

The rich control adds: ``This tool is not the function you are being asked to implement. Do not
pass the task function's parameters into it. Its only parameter, \texttt{code}, is the complete
source text you wrote,'' plus the example \texttt{def add(a,b): return a+b}. Under this warning,
22/100 items still pass the task function's parameters (6/5 discordances, exact $p=1.000$ versus terse). The strict schema
is terse plus \texttt{"strict": true} and \texttt{additionalProperties:false}, with strict tool
calling enabled. This single change takes the wrong-tool count to zero.

\subsection{Thought scaffold and role-disambiguation control}

The Thought-scaffold prefix reads: ``First write your reasoning in a Thought: what the problem
asks, its edge cases, and your intended algorithm. Once you have thought it through, call the
\texttt{run\_tests} tool.'' Nothing else changes; wrong-tool calls rise from 23 to 59
($p<0.001$).

The 1.5B role-disambiguation control adds the following contrast to \texttt{FC\_MANDATORY}:

\begin{Verbatim}[fontsize=\small,breaklines=true,breakanywhere=true]
Common mistake: the function requested by the user is not a callable tool.
The only tool is run_tests, whose sole argument code contains your full source.

Wrong (task asks for join_integers):
{"name":"join_integers","arguments":{"int_array":[1,2,3]}}

Right:
{"name":"run_tests","arguments":{"code":"def join_integers(int_array):\n
    return ','.join(map(str, int_array))"}}
\end{Verbatim}

Recoverable call-shaped outputs rise from 52 to 64, but correctly named calls remain zero and
final pass falls from 15 to 13.

\subsection{Intent criterion}

Emission indicators are benchmark-specific and held fixed across arms within each benchmark.
For KodCode, the regex below targets \texttt{run\_tests} and a quoted code payload with
Python-like lexical markers. It rejects some schema echoes and placeholders but does not
establish JSON validity, Python executability, or semantic intent. Strong and weak use
disjoint trajectory-label sets; tight is computed independently.

For BFCL, the JSON-call indicator detects an embedded decodable JSON object naming a tool
offered in the request and containing object-valued arguments, including arguments encoded
as a JSON string. Format-marker and JSON-marker indicators are independent broader checks;
they overlap and must not be summed. Counts refer to unparsed first-request responses,
with at most one count per response for each indicator.

For $\tau$-bench, the unparsed call-shaped indicator detects a generic named-call pattern
or a Hermes tool-call marker in assistant messages without structured tool calls. It does
not validate membership in the offered tool set or guarantee syntactic validity.
KodCode annotation precision is not assumed to transfer to either external benchmark.

\begin{Verbatim}[fontsize=\small,breaklines=true,breakanywhere=true]
TIGHT = re.compile(
    r'"name"\s*:\s*"run_tests".{0,200}?"arguments"\s*:\s*\{'
    r'.{0,80}?"code"\s*:\s*"(.{0,4000}?)"\s*\}', re.S)
REAL = re.compile(r'\\n|def |class |return |import |lambda ')

def is_tight_call(raw):
    m = TIGHT.search(raw or "")
    return bool(m and REAL.search(m.group(1)))
\end{Verbatim}

The regex requires \texttt{name} before \texttt{arguments} and a closing argument brace
after the quoted code field; valid alternative key orders may be missed.
KodCode human validation is reported in Appendix~\ref{app:humanval}.

\section{Serving and training configuration}
\label{app:config}

\subsection{Per-family serving flags}

\paragraph{Baseline provenance and scope.}
The vLLM 0.27.1 tool-calling documentation, under ``Qwen Models'', lists
\texttt{Qwen/Qwen2.5-*} with the \texttt{hermes} parser
(\url{https://docs.vllm.ai/en/v0.27.1/features/tool_calling/#qwen-models};
checked 23 September 2026). Our Coder baseline applies this family-level guidance;
the passage does not separately validate a Coder checkpoint. Thus ``documented'' denotes
the origin of the configuration choice, not a Coder-specific compatibility guarantee or
evidence of its prevalence in deployment. Issue \#32926 already described this mismatch
before our experiments. Our contribution is its controlled measurement and downstream
consequences, not discovery of the incompatibility or invention of the community repair.

These are the configurations under test, including documented serving flags and the
adapter/schema interventions identified in our experiments. The rightmost column summarizes
observed limitations or failure symptoms; it is not a uniform flag-omission ablation.

{\small
\begin{longtable}[]{@{}p{.17\linewidth}p{.48\linewidth}p{\dimexpr.35\linewidth-4\tabcolsep\relax}@{}}
\caption{Serving configurations and observed symptoms (vLLM 0.27.1).}
\label{tab:serving}\\
\toprule
Family & Flags / intervention & Observed symptom \\
\midrule
\endfirsthead
\toprule
Family & Flags / intervention & Observed symptom \\
\midrule
\endhead
all & \texttt{-{}-enable-auto-tool-choice} & HTTP 400 if omitted \\
Qwen2.5 (Instruct) & \texttt{-{}-tool-call-parser hermes} & 1--88/100 parsed across sizes; unparsed outputs include payload failures \\
Qwen2.5-Coder & Baseline: \texttt{hermes}; repair: community parser plus paired template & baseline 0 parsed across sizes; joint adapter 196/200 on BFCL \\
Llama 3.1 / 3.2 & \texttt{-{}-tool-call-parser llama3\_json} \newline
  \texttt{-{}-chat-template} \path{tool_chat_template_llama3.1_json.jinja} \newline
  plus \texttt{"strict": true} in the schema & malformed arguments; without
  \texttt{strict} the terse baseline has 23\% wrong-tool calls \\
Mistral (HF format) & \texttt{-{}-tokenizer-mode hf -{}-config-format hf -{}-load-format hf}
  \newline \texttt{-{}-tool-call-parser mistral} \newline
  \texttt{-{}-chat-template\allowbreak\ tool\_\allowbreak chat\_\allowbreak template\_\allowbreak mistral\_\allowbreak parallel.jinja} \newline
  \texttt{tool\_call\_id} must be exactly 9 characters & HTTP 400; the officially
  recommended parallel template \emph{raises} the error rate from 2\% to 42\% \\
DeepSeek-Coder & none available: the chat template never injects \texttt{tools} & n/a \\
\bottomrule
\end{longtable}
}

Missing auto-tool-choice support and the tested Mistral configurations produce explicit
request errors. Other mismatches can return HTTP 200 with empty \texttt{tool\_calls},
indistinguishable at that field from declining to call a tool. The released 98-line
\texttt{preflight\_toolcall.py} checks a canonical request for a non-empty call with the
right name and parseable arguments, then repeats under \texttt{tool\_choice: required}
as a positive control. This is a screening check, not a guarantee of compatibility on all tasks.

\subsection{Training configuration}

The historical 1.5B runs use the following configuration; they motivate the investigation but are
not the formal parser-repair control:

{\small
\begin{longtable}[]{@{}p{.27\linewidth}p{\dimexpr.73\linewidth-2\tabcolsep\relax}@{}}
\caption{Training configuration. All arms share every value below; arms differ only in the
protocol and, for the historical runs, the algorithm and seed.}\label{tab:trainconfig}\\
\toprule
Setting & Value \\
\midrule
\endfirsthead
\toprule
Setting & Value \\
\midrule
\endhead
Model & Qwen2.5-Coder-1.5B-Instruct \\
Framework & verl 0.9.0 (vLLM 0.27.1 rollout backend) \\
Algorithm & GRPO ($\times$2 seeds) and RLOO ($\times$1 seed) \\
Trajectories per step & \texttt{train\_batch\_size} 16 $\times$ \texttt{rollout.n} 8 = 128 \\
Steps & 150 for the historical runs and formal 7B control; 10 / 3 / 75 for the historical
diagnostic arms in Appendix~\ref{app:extra} \\
Training data & KodCode, decontaminated pool of 7669 items \\
Held-out evaluation & Multi: 164 HumanEval+, 378 MBPP+; repair: 155 HumanEval+, 299 MBPP+ \\
Denominators & Formal: 542 multi + 454 repair records; historical multi: 540 after excluding two defective items \\
Agent loop & \texttt{tool\_agent} (FC) or \texttt{react\_agent} (ReAct) \\
Tool & \texttt{run\_tests}, single \texttt{code} argument, sandboxed \texttt{pytest} \\
System prompt & \texttt{FC\_MANDATORY} / ReAct mandatory --- matched strength (Appendix~C) \\
Hardware & single A800 80\,GB \\
\bottomrule
\end{longtable}
}

The formal control of \S\ref{sec:rl} instead uses Qwen2.5-Coder-7B-Instruct, GRPO, LoRA rank and
alpha 32 over all linear modules, seed 0, 150 steps, and the same 16 prompts $\times$ 8 rollouts
per step in both arms. Both use the mandatory FC prompt and fixed verifier, retain complete rollout
text, and evaluate the same 542 paired multi-turn items at steps 0/30/60/90/120/150. Evaluation
uses a shared parser-independent programmatic-feedback protocol: the current LoRA is queried
through verl's native rollout RPC at temperature 0, for at most four turns and 1,024 tokens per
turn; generated code is executed against EvalPlus tests and the observation is appended as a user
message. Native RPC metadata must match the expected global weight step. The sole arm
difference is \texttt{multi\_turn.format=hermes} versus the registered
\texttt{qwen2\_5\_coder} parser. Full-parameter 7B RL did not fit the single 80\,GB card, so both
arms use the same parameter-efficient configuration. Logged training settings are learning rate
$10^{-6}$, temperature 1.0, top-$p=1$, top-$k=-1$, prompt budget 2,048 and response budget 6,144
tokens, at most five assistant turns, a 1,200-token tool-response cap and one dispatched tool
call per generation. These are
training budgets, distinct from the four-turn, 1,024-token-per-turn held-out evaluation.
Decontamination uses question n-grams 8/5 and solution n-grams 10/6, with a five-gram minimum;
containment is intersection divided by the smaller set, and the clean-ID selection threshold
is 0.10. These checks do not rule out semantic overlap or pretraining contamination.
This preserves the arm contrast but prevents
a strict comparison to the historical full-parameter curves.

\subsection{Tool-call instrumentation}

verl 0.9.0's step metrics contain \textbf{no} tool-call count. The metric named
\texttt{timing\_s/agent\_loop/tool\_calls} is \emph{elapsed seconds}; reading it as a count is
a hard error and we flag it because the name invites exactly that.

We therefore instrumented the two agent-side call sites directly, each writing one labelled
line per call:

\begin{itemize}
\tightlist
\item \texttt{CodeTool.execute} --- the function-calling path
\item \texttt{ReActAgentLoop.\_run\_tests} --- the ReAct path
\end{itemize}

\textbf{The counter must not be placed in \texttt{sandbox.run\_tests}.} The reward function
evaluates the final program through that same function, so a counter there conflates reward
evaluation with agent tool use. We hit this and moved the counter; the sandbox's own counter
was renamed \texttt{sandbox\_exec\_count} so the two can never be confused again.

In the historical diagnostic run, the FC counter file has \textbf{0 lines}; the ReAct arm's has \textbf{788}, all
carrying the \texttt{react} label, so the provenance is uncontaminated. Three mutually
independent signals agree --- \texttt{num\_turns} pinned at its minimum of 2, tool time
exactly 0.00\,s, and a dedicated counter at 0.

\subsection{Constructing the repaired-FC training arm}

The control initially appeared blocked at three layers:

\begin{enumerate}
\tightlist
\item \texttt{multi\_turn.format=hermes} selects \textbf{verl's own} \texttt{ToolParser}
  registry, which is independent of vLLM's --- so installing a vLLM parser plugin has no
  effect on the training path.
\item A verl-side parser we wrote does recover calls, but \textbf{none of them are the tool}:
  36/52 target the task function, 16/52 carry executable code elsewhere in the body. Name
  normalisation cannot repair this, because no call carries code to normalise.
\item The 1.5B checkpoint emits no correctly named call for that parser to accept, and 7B
  full-parameter training does not fit a single 80\,GB card.
\end{enumerate}

The first two observations determine the intervention; the third determines the scale and LoRA
configuration. We registered the parser inside verl and ran the matched 7B pair. Thus the earlier
statement that no executable repaired-FC arm existed was a description of the configuration
surface, not a permanent limitation; \S\ref{sec:rl} reports the completed control.

\subsection{The path to the repaired-FC training arm}
\label{app:norepair}

We encountered four successive obstacles while constructing the control. Each attempt is a
measurement, so we retain the record; all were eventually resolved by registering the parser in
verl and moving to matched 7B LoRA arms.

\textbf{(i) A vLLM parser plugin does not reach the training loop.} verl's AgentLoop performs its
own tool-call parsing (\texttt{verl/experimental/agent\_loop/tool\_parser.py}, registered as
\texttt{hermes}, adapted from vLLM v0.9.1). \texttt{multi\_turn.format=hermes} selects \emph{that}
parser, not the server's, so attaching a plugin to the vLLM rollout server changes nothing in
training --- a distinction worth stating because the two look identical from the outside.

\textbf{(ii) A verl-side parser recovers calls, but none of them are the tool.} We implemented a
\texttt{qwen2\_5\_coder} parser inside verl's registry accepting \texttt{<tools>} tags,
\texttt{json} fenced blocks and bare JSON. In the historical 1.5B diagnostic with a recorded
mandatory-prompt file override ($n{=}100$), it lifts recoverable calls from
\textbf{0/100 (hermes) to 52/100}. The original override file is unavailable locally,
so exact prompt-byte equivalence to training cannot be verified (Appendix~\ref{app:arms}). Of those 52,
\textbf{zero name} \texttt{run\_tests} (Table~\ref{tab:probe_categories}). Every call targets the
function the task asks the model to \emph{write} --- \texttt{can\_form\_word(\{"tiles", "word"\})},
\texttt{check\_password\_strength(\{"password"\})} --- the same signature as Llama's failure in
\S\ref{sec:llama}, now at a 5$\times$ smaller model. \textbf{Name normalisation cannot repair
this: no call carries code to normalise.}

\textbf{(iii) A role-disambiguation few-shot makes it worse, not better.} Since the failure is
semantic rather than syntactic, we targeted it directly: a system prompt containing the wrong call
and the right call side by side, with an explicit warning that the task function is not a tool.
Same model, same 100 items, same server; only the system prompt changes. The intervention made the
model \emph{more fluent at emitting the wrong call} (52 $\to$ 64 recoverable) and slightly worse at
the task (15 $\to$ 13), with the count that matters staying at zero
(Table~\ref{tab:pressure}). This is the third independent instance of the pattern in
\S\ref{sec:pressure}.

\textbf{(iv) Constrained decoding is unavailable in this stack.} Forcing the tool choice, or
schema-constrained decoding, is not exposed by verl 0.9.0's vLLM rollout path (its only
\texttt{strict} key belongs to the profiler). This is a limitation of the tested rollout path, not an argument against constrained decoding.
Parser repair accepts existing emissions, whereas generation constraints alter the output
distribution. The latter is a distinct potential remedy for serialization or allowed-name
violations; neither remedy by itself establishes correct arguments or successful task behavior.

\section{Arm inventory}
\label{app:arms}

Every full-length arm, generated from the trajectory files' own provenance rather than
transcribed by hand. All 50 arms share one item set (\texttt{clean[:100]}; distinct-set
count verified = 1), so every cross-arm comparison in the paper is exactly paired. Fields
recorded as \texttt{---} were absent from that generation's provenance schema
(Appendix~A.4); this is disclosed rather than back-filled. Seven arms carry a known-wrong
\texttt{max\_tokens} record and are listed in Appendix~A.1; five are inadmissible for pass
rates and are listed in Appendix~A.3.

The \texttt{strength} column is the recorded CLI selector, not necessarily the effective prompt.
For \texttt{a14} and \texttt{roleprobe}, records specify \texttt{sys\_file}; the archived probe
loads this file after selecting the built-in prompt, overriding \texttt{strength=optional}.
The retained role-disambiguation file explicitly mandates \texttt{run\_tests}.
The original \texttt{sys\_fc\_mandatory.txt} file was not found in the local archive, so the
a14 run's exact prompt bytes cannot be independently verified; its name alone is not proof.
The zero correctly named calls remain an observed outcome, not a fully byte-matched prompt replication.

Not listed: \texttt{v8\_Llama8B\_recheck} (23 items, a targeted qualitative re-run) and
eight \texttt{*smoke*} files at n=3. Neither is a formal arm.

{\scriptsize
\setlength{\tabcolsep}{2.5pt}
\begin{longtable}[]{@{}llllllrrr@{}}
\caption{All 50 full-length arms and their recorded configuration.}\label{tab:arms}\\
\toprule
Arm & Model & Prot. & Strength & Schema & Adapter & tok & temp & seed \\
\midrule
\endfirsthead
\toprule
Arm & Model & Prot. & Strength & Schema & Adapter & tok & temp & seed \\
\midrule
\endhead
a14\_Qwen1.5B\_fc\_mandatory & Qwen-1.5B & fc & optional & terse & cross\_family & 2048 & --- & --- \\
roleprobe\_Qwen1.5B\_fc\_roledisambig & Qwen-1.5B & fc & optional & terse & cross\_family & 2048 & 0.0 & 0 \\
v11\_DS1.3b\_react & DS-1.3B & react & optional & terse & cross\_family & 2048 & --- & --- \\
v11\_DS6.7b\_react & DS-6.7B & react & optional & terse & cross\_family & 2048 & 0.0 & --- \\
v11\_Llama32\_1B\_fc\_strict & Llama-3.2-1B & fc & optional & strict & cross\_family & 2048 & 0.0 & --- \\
v11\_Llama32\_1B\_react & Llama-3.2-1B & react & optional & terse & cross\_family & 2048 & 0.0 & --- \\
v11\_Llama32\_3B\_fc\_strict & Llama-3.2-3B & fc & optional & strict & cross\_family & 2048 & 0.0 & --- \\
v11\_Llama32\_3B\_react & Llama-3.2-3B & react & optional & terse & cross\_family & 2048 & 0.0 & --- \\
v13\_DS1.3b\_react & DS-1.3B & react & optional & terse & cross\_family & 2048 & 0.0 & --- \\
v13\_DS6.7b\_react & DS-6.7B & react & optional & terse & cross\_family & 2048 & 0.0 & --- \\
v13\_Llama32\_1B\_react & Llama-3.2-1B & react & optional & terse & cross\_family & 2048 & 0.0 & --- \\
v13\_Llama32\_3B\_react & Llama-3.2-3B & react & optional & terse & cross\_family & 2048 & 0.0 & --- \\
v13\_Llama8B\_fcstrict\_t06\_s1 & Llama-3.1-8B-Instruct & fc & optional & strict & cross\_family & 2048 & 0.6 & 1 \\
v13\_Llama8B\_fcstrict\_t06\_s2 & Llama-3.1-8B-Instruct & fc & optional & strict & cross\_family & 2048 & 0.6 & 2 \\
v13\_Llama8B\_fcstrict\_t06\_s3 & Llama-3.1-8B-Instruct & fc & optional & strict & cross\_family & 2048 & 0.6 & 3 \\
v13\_Llama8B\_react\_t06\_s1 & Llama-3.1-8B-Instruct & react & optional & terse & cross\_family & 2048 & 0.6 & 1 \\
v13\_Llama8B\_react\_t06\_s2 & Llama-3.1-8B-Instruct & react & optional & terse & cross\_family & 2048 & 0.6 & 2 \\
v13\_Llama8B\_react\_t06\_s3 & Llama-3.1-8B-Instruct & react & optional & terse & cross\_family & 2048 & 0.6 & 3 \\
v14\_Llama8B\_react & Llama-3.1-8B-Instruct & react & optional & terse & cross\_family & 2048 & 0.0 & --- \\
v14\_Mistral7B\_react & Mistral-7B & react & optional & terse & cross\_family & 2048 & 0.0 & --- \\
v14\_Qwen7B\_react & Qwen-7B & react & optional & terse & cross\_family & 2048 & 0.0 & --- \\
v3\_Llama8B\_fc & Llama-3.1-8B-Instruct & fc & optional & --- & cross\_family & 2048 & --- & --- \\
v3\_Llama8B\_fc\_legacy & Llama-3.1-8B-Instruct & fc & optional & --- & cross\_family & 2048 & --- & --- \\
v3\_Llama8B\_react & Llama-3.1-8B-Instruct & react & optional & --- & cross\_family & 2048 & --- & --- \\
v3\_Mistral7B\_fc & Mistral-7B & fc & optional & terse & cross\_family & 2048 & --- & --- \\
v3\_Mistral7B\_react & Mistral-7B & react & optional & terse & cross\_family & 2048 & --- & --- \\
v3\_Qwen7B\_fc\_legacy & Qwen-7B & fc & optional & terse & cross\_family & 2048 & --- & --- \\
v3\_Qwen7B\_fc\_nosuffix & Qwen-7B & fc & optional & terse & cross\_family & 2048 & --- & --- \\
v3\_Qwen7B\_react & Qwen-7B & react & optional & terse & cross\_family & 2048 & --- & --- \\
v4\_Llama8B\_fc\_cot & Llama-3.1-8B-Instruct & fc & optional & terse & cross\_family & 2048 & --- & --- \\
v4\_Llama8B\_fc\_rich & Llama-3.1-8B-Instruct & fc & optional & rich & cross\_family & 2048 & --- & --- \\
v5\_Llama8B\_fc\_official & Llama-3.1-8B-Instruct & fc & optional & terse & cross\_family & 2048 & --- & --- \\
v5\_Mistral7B\_fc\_official & Mistral-7B & fc & optional & terse & cross\_family & 2048 & --- & --- \\
v5\_Qwen1.5B\_fc\_intent & Qwen-1.5B & fc & optional & terse & cross\_family & 2048 & --- & --- \\
v5\_Qwen14B\_fc\_intent & Qwen-14B & fc & optional & terse & cross\_family & 2048 & --- & --- \\
v5\_Qwen32B\_fc\_intent & Qwen-32B & fc & optional & terse & cross\_family & 2048 & --- & --- \\
v5\_Qwen3B\_fc\_intent & Qwen-3B & fc & optional & terse & cross\_family & 2048 & --- & --- \\
v5\_Qwen7B\_fc\_intent & Qwen-7B & fc & optional & terse & cross\_family & 2048 & --- & --- \\
v6\_Llama8B\_fc\_strict & Llama-3.1-8B-Instruct & fc & optional & strict & cross\_family & 2048 & --- & --- \\
v6b\_Qwen1.5B\_fc\_plugin & Qwen-1.5B & fc & optional & terse & cross\_family & 2048 & --- & --- \\
v6b\_Qwen3B\_fc\_plugin & Qwen-3B & fc & optional & terse & cross\_family & 2048 & --- & --- \\
v6b\_Qwen7B\_fc\_plugin & Qwen-7B & fc & optional & terse & cross\_family & 2048 & --- & --- \\
v9\_Mistral7B\_fc\_official\_strict & Mistral-7B & fc & optional & strict & cross\_family & 2048 & --- & --- \\
v9\_Mistral7B\_fc\_strict & Mistral-7B & fc & optional & strict & cross\_family & 2048 & --- & --- \\
x2\_DS1.3B\_react & DS-1.3B & react & optional & --- & --- & --- & --- & --- \\
x2\_DS6.7B\_react & DS-6.7B & react & optional & --- & --- & --- & --- & --- \\
x2\_Llama8B\_fc & Llama-3.1-8B-Instruct & fc & optional & --- & --- & --- & --- & --- \\
x2\_Llama8B\_react & Llama-3.1-8B-Instruct & react & optional & --- & --- & --- & --- & --- \\
x2\_Mistral7B\_fc & Mistral-7B & fc & optional & --- & --- & --- & --- & --- \\
x2\_Mistral7B\_react & Mistral-7B & react & optional & --- & --- & --- & --- & --- \\
\bottomrule
\end{longtable}
}

\section{Human validation of the intent criterion, in full}
\label{app:humanval}

The scale trend in \S\ref{sec:scale} rests on an automated classifier, so we validated it
against human judgement. 98 outputs were sampled \textbf{stratified by classifier verdict}
(40 \texttt{tight}, 28 \texttt{strong}-but-not-\texttt{tight}, 30 neither) --- deliberately
over-sampling the decision boundary so both false positives and false negatives are estimable.
Sampling is seeded and the annotator saw only the raw output, never the classifier's label.
We report \textbf{two rounds}, because the first is itself a finding.

\begin{center}
\begin{minipage}{\linewidth}
\centering
\small
\captionof{table}{Human validation of the intent criterion, round 1 and its adjudication. The classifier is used only to count what the server discarded, so its recall matters more than its precision.}\label{tab:validation}
\begin{tabular}{@{}lllllll@{}}
\toprule
& agreement & Cohen's $\kappa$ & precision & recall & FP & FN \\
\midrule
Round 1 (raw output only) & 86.7\% & 0.713 & 70.0\% & 96.6\% & 12 & 1 \\
After adjudication & \textbf{96.9\%} & \textbf{0.936} & \textbf{95.0\%} & 97.4\% & 2 & 1 \\
\bottomrule
\end{tabular}
\end{minipage}
\end{center}

The 12 round-1 false positives split into two causes, and the second is ours, not the
annotator's. \textbf{Eleven are reading-order failures}: the tool call sits mid-document, after
a fenced \verb|```python| block (match positions 1209--2585 in outputs of median length
${\sim}2000$); the annotator read the code block, concluded ``this is a direct answer, not a
call'', and stopped. \textbf{One is an instrument failure}: for a 32B item the matched span
begins at character 2585 and the annotation pack truncated every output at 2400, so the
evidence was \emph{not in the pack at all} and that item could not have been labelled
correctly. Re-shown the matched span alone, 10 of 13 disputed items were reversed. The
corrected pack carries every output in full.

\textbf{The first adjudication was one-sided, and we do not treat its $\kappa$ as a reliability
estimate.} Auditing the procedure after the fact: the 13 adjudicated items were selected
\emph{exactly} as the items where the annotator disagreed with the classifier (13/13 overlap),
and all 10 reversals moved \emph{toward} the classifier, none away. Under that selection rule
agreement can only rise, so $0.713 \to 0.936$ is a property of which items were re-examined,
not evidence of reliability. We report the second annotator instead.

\subsection*{A second independent annotator}

A second annotator, blind to the classifier, to the first annotator's labels and to the first
pack's item order, labelled the identical 98 items under a verbatim-identical rubric. Two
changes were made to the instrument: outputs are carried \textbf{in full}, and the adjudication
rule was fixed in advance to review \emph{all} three-way disagreements rather than only those
disagreeing with the classifier.

\begin{center}
\begin{minipage}{\linewidth}
\centering
\small
\captionof{table}{Inter-annotator reliability. Of 98 sampled items, 97 have complete paired A1/A2 records. The identical-byte subset removes a presentation confound introduced by the first pack's 2,400-character cap.}\label{tab:interannotator}
\begin{tabular}{@{}lcc@{}}
\toprule
Cohen's $\kappa$ (95\% CI, bootstrap) & all items ($n$=97) & \textbf{identical bytes ($n$=62)} \\
\midrule
A1 vs.\ A2 (inter-annotator) & 0.736 [0.595, 0.866] & \textbf{0.967 [0.893, 1.000]} \\
A2 vs.\ classifier           & 0.936 [0.852, 1.000] & \textbf{0.967 [0.896, 1.000]} \\
A1 vs.\ classifier           & 0.712 [0.564, 0.849] & 0.934 [0.829, 1.000] \\
\bottomrule
\end{tabular}
\end{minipage}
\end{center}

Of the 12 paired disagreements, 11 have full text exceeding 2,400 characters and A1 saw a
different presentation from A2; the first-round audit attributed most errors to reading order and
one to evidence actually omitted by truncation. We therefore do not assign all 11 to one cause.
Repeated A1--A2 intervals use 5,000 bootstrap resamples, seed 7 and sorted item IDs;
earlier differing bounds reflected inconsistent bootstrap inputs/order, not different annotators.
On the 62 items where both read identical bytes the two agree on 61, and
\textbf{inter-annotator $\kappa$ = 0.967}.

\textbf{Third-party adjudication, and the final labels.} Fifteen items were not unanimous across
A1, A2 and the classifier. All fifteen went to a third adjudicator under the pre-registered rule
--- selected by \emph{any} three-way disagreement, never by disagreement with the classifier ---
who saw full text and no one else's labels. The adjudicator sided with A2 on 11, with the
classifier on 9, and with A1 on 6, and \textbf{moved away from the classifier on 6 of 15}. That
last number is the check that matters: the first round's adjudication moved away from the
classifier zero times out of thirteen, which is what an outcome looks like when the selection
rule has already decided it. Combining the 83 unanimous items with the 15 adjudicated ones gives
a 98-item gold standard. The third adjudicator was blind to classifier and annotator labels; the
classifier nevertheless affected the original stratified sample and which non-unanimous items
required adjudication, so we do not call the gold set classifier-independent.

\begin{center}
\begin{minipage}{\linewidth}
\centering
\small
\captionof{table}{Validation against the adjudicated gold standard. Pairwise denominators differ because one sampled item lacks a complete A1/A2 pair; classifier--gold uses all 98 items.}\label{tab:validation2}
\begin{tabular}{@{}lccc@{}}
\toprule
Cohen's $\kappa$ (95\% CI, bootstrap) & full $n$ & full sample & identical bytes ($n$=62) \\
\midrule
A1 vs.\ A2 (inter-annotator) & 97 & 0.736 [0.595, 0.866] & \textbf{0.967 [0.893, 1.000]} \\
Classifier vs.\ gold standard & 98 & \textbf{0.871 [0.756, 0.958]} & 0.934 [0.829, 1.000] \\
\bottomrule
\end{tabular}
\end{minipage}
\end{center}

\textbf{Sensitivity estimate, not a corrected population total.} Tight-stratum precision is
36/40=0.900. If this precision transfers to the 32B arm, multiplying 80 by 0.900 gives
approximately 72 classifier-positive outputs expected to satisfy the gold criterion. This does
not restore false negatives and is not an estimate of all true calls. The stratified gold sample
contains 4 false positives and 2 false negatives; their unweighted difference is not a population
bias estimate. Uniformly applying 0.900 yields $0,3.6,18.9,32.4,72$ by construction, but
scale-specific precision is unmeasured, so this calculation does not independently validate the trend.

For the record, this number moved twice as the validation improved, and we report the trajectory
rather than only its endpoint: \textbf{0.950} under the first, one-sided adjudication;
\textbf{0.975} under A2's labels alone; \textbf{0.900} against the adjudicated gold standard. The
first was inflated by its selection rule, the second rested on a single rater, and only the third
is built from a rule fixed before the labels were seen.

\textbf{A bound on the measurement itself.} The probe stores at most 4000 characters of raw
output per item; 10 of the 98 sampled outputs reach that ceiling, and one grey-zone item is a
call truncated by it. The classifier and both annotators read identical bytes, which is what
agreement requires, but a call emitted beyond character 4000 is invisible to all three.
All ten cap hits occur in the 32B subset of this stratified sample: by increasing scale the
hit counts are 0/6, 0/13, 0/15, 0/24, and 10/40. These are sample diagnostics, not population
truncation rates, but they show that the uncertainty is concentrated at the headline rung.
The cap therefore creates downward pressure through false negatives. Because the classifier also
has false positives (tight precision 0.900), however, the net bias of the raw count is not known;
we report both the uncorrected count and the assumption-dependent sensitivity estimate rather than calling
the total a strict lower bound.

\section{Supplementary results}
\label{app:extra}

\subsection{Historical observations that motivated the study}

Historical 1.5B runs showed numerical pass gains of 2.6--2.8 points ($p=0.064/0.078/0.077$),
not statistically significant. Of newly passing items, 91--94\% passed on turn 1; multi-turn
rescues ranged from 5 to 10 out of 540 across the reported checkpoint sequences. These small exploratory counts prompted the investigation
but do not identify an interface effect or independently establish absence of learning.

\begin{figure}[htbp]
\centering
\includegraphics[width=\linewidth]{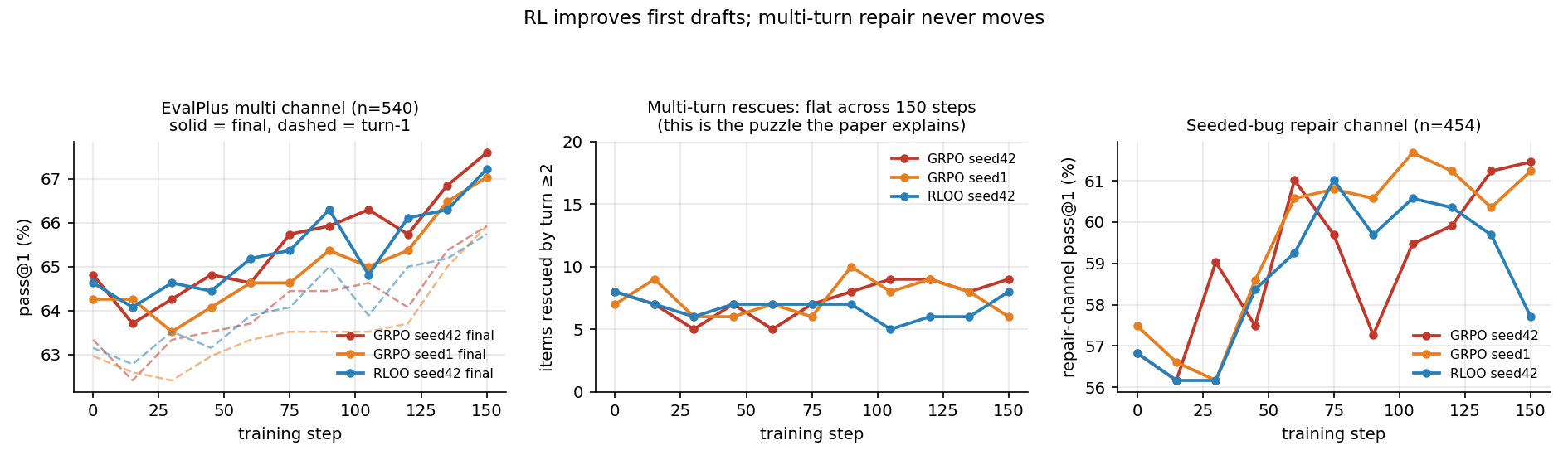}
\caption{Left: pass@1 rises over 150 steps on all three runs, driven by first drafts (dashed)
rather than final answers. Centre: multi-turn rescues range from 5--10 out of 540 across checkpoints, algorithms and seeds. Right:
the seeded-bug repair channel does improve, confirming the model is learning something about
fixing code; no clear increase in multi-turn rescue counts was detected in these historical runs. \S\ref{sec:rl} shows the tool was
never successfully called in any of these runs.}
\label{fig:training_curves}
\end{figure}

\subsection{Result tables referenced from \S\ref{sec:results}}
\label{app:tables}

\begin{center}
\begin{minipage}{\linewidth}
\centering
\small
\captionof{table}{BFCL v4, first HTTP request per case. Emission indicators are measured only where parsing failed; JSON-call, format-marker and JSON-marker checks overlap and must not be summed.}\label{tab:bfclfunnel}
\setlength{\tabcolsep}{3.5pt}\begin{tabular}{@{}llllllll@{}}
\toprule
Arm & cases & HTTP err & first parsed & unparsed & JSON-call & format & JSON \\
\midrule
\texttt{hermes} (documented) & 200 & 0 & \textbf{0} & 200 & 166 & 169 & 187 \\
\texttt{qwen2\_5\_coder} (repaired) & 200 & 0 & \textbf{196} & 4 & 0 & 1 & 2 \\
\bottomrule
\end{tabular}
\end{minipage}
\end{center}
\begin{center}
\begin{minipage}{\linewidth}
\centering
\small
\captionof{table}{$\tau$-bench \texttt{retail}, all 115 tasks paired across arms (none dropped).}\label{tab:tau}
\begin{tabular}{@{}lrr@{}}
\toprule
 & documented & repaired \\
\midrule
assistant turns & 1389 & 1971 \\
server-parsed calls & \textbf{0} & \textbf{636} \\
emitted but unparsed & \textbf{789} & 6 \\
tool executions / observations returned & \textbf{0} & \textbf{636} \\
\midrule
tasks entering the tool loop at all & \textbf{0} & \textbf{103} \\
tasks solved & 7 & 10 \\
\bottomrule
\end{tabular}
\end{minipage}
\end{center}
\begin{center}
\begin{minipage}{\linewidth}
\centering
\small
\captionof{table}{Same probe, items and criterion under a \emph{matched} envelope.}\label{tab:counterfactual}
\begin{tabular}{@{}lll@{}}
\toprule
Size & server-parsed & \textbf{silent count / 100 (tight, unparsed)} \\
\midrule
Qwen3-0.6B & 86 & \textbf{1} \\
Qwen3-1.7B & 44 & \textbf{1} \\
Qwen3-4B & 70 & \textbf{2} \\
Qwen3-8B & 97 & \textbf{0} \\
\midrule
Qwen2.5-Coder 1.5B--32B (mismatched) & 0 at every size & \textbf{0 $\to$ 4 $\to$ 21 $\to$ 36 $\to$ 80} \\
\bottomrule
\end{tabular}
\end{minipage}
\end{center}
\begin{center}
\begin{minipage}{\linewidth}
\centering
\small
\captionof{table}{Qwen2.5-Instruct under the identical mismatched configuration, with both ladders' task outcomes measured on a working executor.}\label{tab:instruct}\label{tab:turn1}
\setlength{\tabcolsep}{3.5pt}\begin{tabular}{@{}llll|ll|lll@{}}
\toprule
& \multicolumn{3}{c|}{Instruct: parsed / unparsed indicators} & \multicolumn{2}{c|}{Coder} & \multicolumn{3}{c}{Instruct} \\
Size & parsed & \textbf{tight} & weak & parsed & turn-1 = final & turn-1 & final & \textbf{rescued} \\
\midrule
1.5B & 1 & 0 & 7 & 0 & 31 & 13 & 13 & \textbf{0} \\
3B & 11 & 56 & 2 & 0 & 37 & 14 & 17 & \textbf{+3} \\
7B & 63 & 21 & 0 & 0 & 53 & 34 & 40 & \textbf{+6} \\
14B & 77 & 19 & 0 & 0 & 53 & 38 & 52 & \textbf{+14} \\
\textbf{32B} & \textbf{88} & 8 & 0 & 0 & 68 & 42 & 65 & \textbf{+23} \\
\bottomrule
\end{tabular}
\end{minipage}
\end{center}
\begin{center}
\begin{minipage}{\linewidth}
\centering
\small
\captionof{table}{Llama-3.1-8B arms, paired on the same 100 items.}\label{tab:llama_arms}
\begin{tabular}{@{}llllll@{}}
\toprule
Arm & Calls issued & Wrong-tool & Executed & Turn-1 & Final \\
\midrule
terse (parser only) & 97 & 23 & 74 & 34 & 49 \\
+ official template & 95 & 22 & 73 & 33 & 44 \\
+ official template \textbf{+ \texttt{strict}} & 98 & \textbf{0} & \textbf{97} & \textbf{46} & \textbf{61} \\
ReAct (reference) & 98 & 0 & 97 & \textbf{61} & 80 \\
\bottomrule
\end{tabular}
\end{minipage}
\end{center}
\begin{center}
\begin{minipage}{\linewidth}
\centering
\small
\captionof{table}{Decomposing the ReAct--FC gap on Llama, paired McNemar.}\label{tab:decomposition}
\begin{tabular}{@{}llll@{}}
\toprule
Comparison & b/c & \emph{p} & Descriptive difference \\
\midrule
terse-FC $\to$ strict-FC (49 $\to$ 61) & 15/3 & 0.0075 & \textbf{interface repair, +12} \\
strict-FC $\to$ ReAct (61 $\to$ 80) & 27/8 & 0.0019 & \textbf{protocol, +19} \\
terse-FC $\to$ ReAct (49 $\to$ 80) & 38/7 & 3.1e-06 & both, +31 \\
\bottomrule
\end{tabular}
\end{minipage}
\end{center}
\begin{center}
\begin{minipage}{\linewidth}
\centering
\small
\captionof{table}{Repair loop on Qwen2.5-Coder-7B. Tested steps count generations with extracted code sent to the sandbox, not all assistant outputs. The historical ReAct row uses v3; the unified v14 rerun has mean tested steps 1.93 with unchanged pass/rescue counts.}\label{tab:repair}
\begin{tabular}{@{}lllllll@{}}
\toprule
Arm & Parsed & Mean tested steps & $\geq$2 tested steps & Turn-1 & Final & Rescued \\
\midrule
ReAct & 95 & 1.92 & 40 & 57 & \textbf{74} & 17 \\
FC + hermes & \textbf{0} & 0.96 & \textbf{0} & 53 & 53 & \textbf{0} \\
FC + dedicated adapter & \textbf{84} & 1.82 & \textbf{37} & 53 & \textbf{62} & \textbf{9} \\
\bottomrule
\end{tabular}
\end{minipage}
\end{center}
\begin{center}
\begin{minipage}{\linewidth}
\centering
\small
\captionof{table}{RL training rollouts, matched in model, data, hyper-parameters, seed and prompt strength.}\label{tab:training}
\begin{tabular}{@{}lllll@{}}
\toprule
Arm & Steps & \texttt{num\_turns/mean} & Tool time & Tool calls / rollout \\
\midrule
FC (\texttt{tool\_agent} + hermes) & 10 & \textbf{2.000} (constant) & \textbf{0.00 s} & \textbf{0.000} \\
ReAct (\texttt{react\_agent}) & 3 & 5.883 & 12.21 s & \textbf{2.052} \\
\bottomrule
\end{tabular}
\end{minipage}
\end{center}

\subsection{Offline re-parse matrix and failure-layer decomposition}
\label{app:reparse}

To exclude ``your extractor is simply better than \texttt{hermes}'', we re-parsed the
\emph{same stored bytes} under four rules (\texttt{analysis/reparse\_matrix.py}, no GPU).

\begin{center}
\begin{minipage}{\linewidth}
\centering
\small
\captionof{table}{The same bytes, re-parsed offline under four rules. Where the server already parsed a call, \texttt{content} is empty and offline replay has nothing to read; those cells are N/A rather than zero. The gap is a parser mismatch, not a claim that no parser could have read the output.}\label{tab:reparse}
\begin{tabular}{@{}llllll@{}}
\toprule
Arm (server-parsed = 0) & server & hermes replay & \texttt{<tools>} replay & bare JSON & tight \\
\midrule
Qwen-1.5B & 0 & 0 & 0 & 0 & 0 \\
Qwen-3B & 0 & 0 & 0 & 5 & 4 \\
Qwen-7B & 0 & 0 & 0 & 22 & 21 \\
Qwen-14B & 0 & 0 & 0 & 42 & 36 \\
Qwen-32B & 0 & \textbf{0} & 0 & \textbf{100} & \textbf{80} \\
\bottomrule
\end{tabular}
\end{minipage}
\end{center}

The \texttt{hermes} column is zero at every scale: the model never produces that format. The
bare-JSON and tight columns are not: the calls exist, and their difference from the server
column is precisely what the format mismatch removes. For arms where the server \emph{did}
parse, offline content-only re-parsing is \textbf{undefined, not zero} --- a successful parse
moves the call into \texttt{tool\_calls} and leaves \texttt{content} empty.

Replaying vLLM~0.27.1's \texttt{hermes} extractor line for line over the stored bytes
(\texttt{analysis/failure\_layer.py}) separates the layer at which each item is lost: no
\texttt{<tool\_call>} envelope at all; envelope present but the JSON payload malformed; payload
well-formed and rejected only because \texttt{json.loads} does not tolerate trailing text in the
capture; or a genuine parser loss, meaning the replay succeeds where the server did not.

\begin{center}
\begin{minipage}{\linewidth}
\centering
\small
\captionof{table}{Where the unparsed items are lost. These are first-failure categories of all unparsed outputs, not counts of identified tool attempts. \textbf{Genuine parser loss is zero in every arm} --- and across all 4254 de-duplicated first turns in the archive.}\label{tab:layer}
\begin{tabular}{@{}lllll@{}}
\toprule
Arm & no envelope & payload malformed & strictness only & \textbf{parser loss} \\
\midrule
Coder 1.5B--32B (each size) & 100 & 0 & 0 & \textbf{0} \\
Instruct 1.5B & 93 & 6 & 0 & \textbf{0} \\
Instruct 3B & 22 & 41 & 26 & \textbf{0} \\
Instruct 7B & 3 & 33 & 1 & \textbf{0} \\
Instruct 14B & 13 & 6 & 4 & \textbf{0} \\
Instruct 32B & 1 & 9 & 2 & \textbf{0} \\
\bottomrule
\end{tabular}
\end{minipage}
\end{center}

Missing envelopes alone do not identify censoring: the 100 Coder outputs at each size include
both call-shaped and non-call outputs (notably at 1.5B). A separate emission criterion is
needed to identify candidate masking. Instruct outputs reaching the payload check may break JSON, almost always by embedding Python whose docstring quotes or
raw newlines are left unescaped inside the \texttt{code} string. And \textbf{in no arm --- across
the whole archive --- did the server fail to parse a call that its own extractor would have
accepted.} ``Censoring'' throughout this paper therefore means \emph{the attempt never reaches
the parser in a form it is defined to accept}, not \emph{the parser discards valid calls}.

The repair-loop table uses the stored \texttt{n\_turns} counter, which advances only when code
is extracted and tested. The Hermes arm has 100 first assistant outputs but four yield no code,
hence 96/100 tested steps. Its mean assistant-output count is 1.00; the adapter's is 1.90,
rather than its 1.82 tested-step mean. Reaching a second tested step is stricter than merely
receiving feedback: a subsequent unextractable response does not increment this counter.

\subsection{BFCL: gating and the full reproduction record}
\label{app:bfcl}

\paragraph{Fixed-output replay on single-turn cases.}
\label{app:fixedreplay}
We added a post-hoc, answer-blind replay of all 100 \texttt{simple\_python} documented-arm
responses from the clean run. A deterministic extractor scans left to right for maximal JSON
objects or arrays, accepts named calls with object arguments (including JSON-encoded argument
strings), and retains all recovered calls. It neither repairs malformed JSON nor selects a
call using the answer key. All 97 recovered responses name an offered tool and pass structural
signature checks (required fields, types, declared parameter names, array items and enums).
Format annotations are not validated. Of the 100 original responses, 92 pass the official
BFCL AST checker at the experiment's pinned commit. The 100 repaired-arm outputs reproduce
96 official successes as a positive control. Scoring code is unchanged; the offline loader
supplies the same \texttt{underscore\_to\_dot=True} registry property without importing
online inference SDKs.

This is an additional offline interpretation rule, not the community dedicated parser:
none of the original 100 responses contains its required \texttt{<tools>} start tag.
The result demonstrates that correct single-turn calls can be recovered from fixed emitted
bytes. It does not attribute the remaining difference to template changes alone, establish
a safe general-purpose fallback parser, or recover a counterfactual multi-turn success rate.
Changing a multi-turn action would change subsequent environment feedback and generations.
The supplement includes raw responses, the recorded request schemas, answer keys, per-item
verdicts and the pinned checker sources; see Appendix~\ref{app:repro}.

Each arm passes a two-case end-to-end smoke test before its full run, and a run is rejected
unless the result count, the score file's \texttt{total\_count} and the number of distinct case
ids in the recording proxy all agree at 200 with zero HTTP errors. Both arms run at \texttt{max\_model\_len = 16384}, \texttt{temperature = 0} and inference seed
0, on cases drawn at manifest seed 20260901. Registration of BFCL's OpenAI
handler happens in memory at import time, so \texttt{bfcl\_eval}'s own files are left
byte-identical; a recording proxy sits between benchmark and server and logs every request.

We ran the pair twice. The \texttt{hermes} arm is identical across runs in every column: 0/200
parsed, 166 tight, 480 requests, 0 errors. The repaired arm moves by one case between runs ---
\textbf{197/200 versus 196/200} first parsed, \textbf{99/100 versus 98/100} executed --- while success is 19 in
both. Greedy decoding does not make vLLM bitwise deterministic across differing batch
compositions, and this arm runs eight threads through a multi-turn executor, so we report its
intermediate funnel columns as $\pm1$ rather than exact.

\begin{center}
\begin{minipage}{\linewidth}
\centering
\small
\captionof{table}{BFCL \texttt{multi\_turn\_base}, n = 100, joined to the HTTP trace by case id.}\label{tab:bfcl5}
\begin{tabular}{@{}lllllll@{}}
\toprule
Arm & tight (unparsed) & first parsed & any parsed & executed & \textbf{success} & rescued \\
\midrule
\texttt{hermes} & 69 & 0 & 0 & 0 & \textbf{0} & 0 \\
\texttt{qwen2\_5\_coder} & 0 & 97 & 98 & 98 & \textbf{19} & 0 \\
\bottomrule
\end{tabular}
\end{minipage}
\end{center}

\begin{center}
\begin{minipage}{\linewidth}
\centering
\small
\captionof{table}{Two quantified failure layers; the full four-family table is Table~\ref{tab:taxonomy_full}.}\label{tab:taxonomy}
\begin{tabular}{@{}p{.14\linewidth}p{.10\linewidth}p{.25\linewidth}p{.22\linewidth}p{\dimexpr.29\linewidth-8\tabcolsep\relax}@{}}
\toprule
Family & Layer & Symptom & Detectable in advance? & Remedy \\
\midrule
Qwen2.5-Coder & parser & emits a fenced JSON block, not \texttt{<tool\_call>}
  & $\times$ template check \textbf{false-positives} & dedicated adapter \\
Llama-3.1-8B & schema & calls the \emph{task function} as a tool (23\%)
  & $\times$ requires argument inspection & \texttt{strict: true} \\
\bottomrule
\end{tabular}
\end{minipage}
\end{center}

\subsection{The full four-family taxonomy}
\label{app:taxonomy}

\begin{center}
\begin{minipage}{\linewidth}
\centering
\small
\captionof{table}{Four serving diagnostics, not an exhaustive taxonomy. Only Mistral's failure announces itself.}\label{tab:taxonomy_full}
\begin{tabular}{@{}p{.14\linewidth}p{.10\linewidth}p{.25\linewidth}p{.22\linewidth}p{\dimexpr.29\linewidth-8\tabcolsep\relax}@{}}
\toprule
Family & Layer & Symptom & Detectable in advance? & Remedy \\
\midrule
DeepSeek-Coder & template & chat template never injects tools
  & $\checkmark$ template inspection & none \\
Qwen2.5-Coder & parser & emits a fenced JSON block, not \texttt{<tool\_call>}
  & $\times$ template check \textbf{false-positives} & dedicated adapter \\
Llama-3.1-8B & schema & calls the \emph{task function} as a tool (23\%)
  & $\times$ requires argument inspection & \texttt{strict: true} \\
Mistral-7B-v0.3 & token & repeats \texttt{[TOOL\_CALLS]} $\to$ HTTP 400
  & $\sim$ errors, but only sometimes & \textbf{none found} \\
\bottomrule
\end{tabular}
\end{minipage}
\end{center}

\begin{figure}[htbp]
\centering
\includegraphics[width=\linewidth]{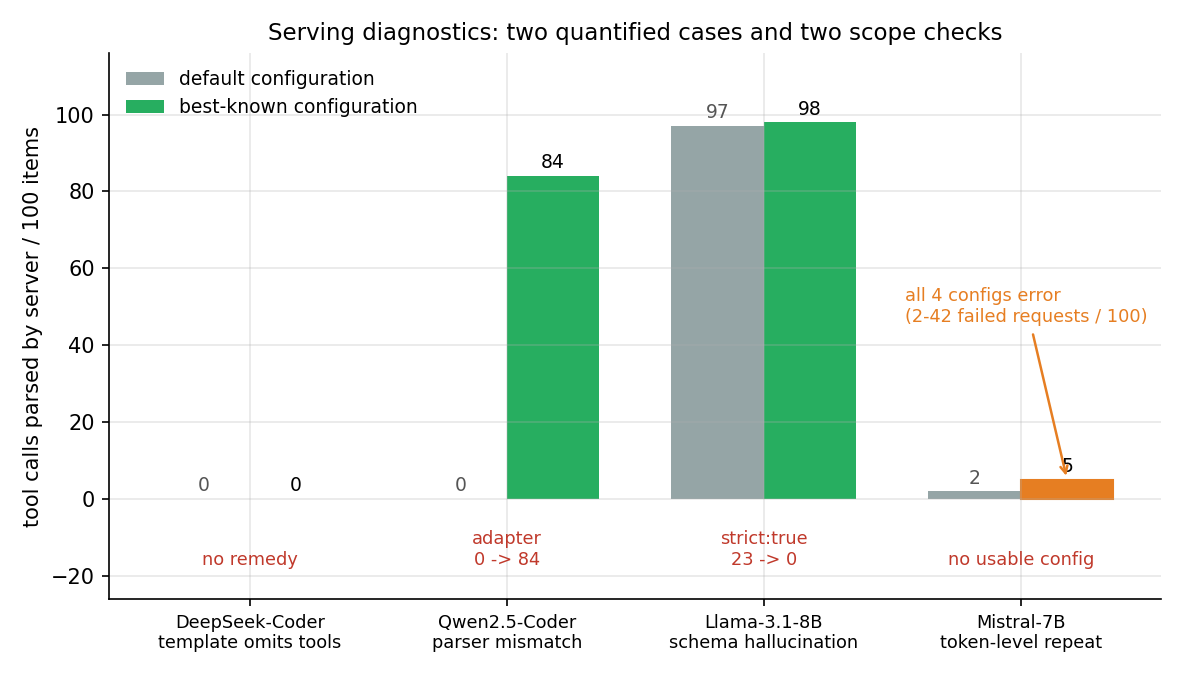}
\caption{Serving diagnostics with a consistent server-parsed metric. Llama is 97 versus 98 parsed; wrong-tool and executed counts are reported separately. Grey is the configuration a reader arrives at by
following the documentation; green is the best configuration we found. Mistral never produces an
error-free run under any of four configurations, so its bars are not comparable with the others
and are drawn in orange.}
\label{fig:taxonomy}
\end{figure}

\begin{figure}[htbp]
\centering
\includegraphics[width=0.72\linewidth]{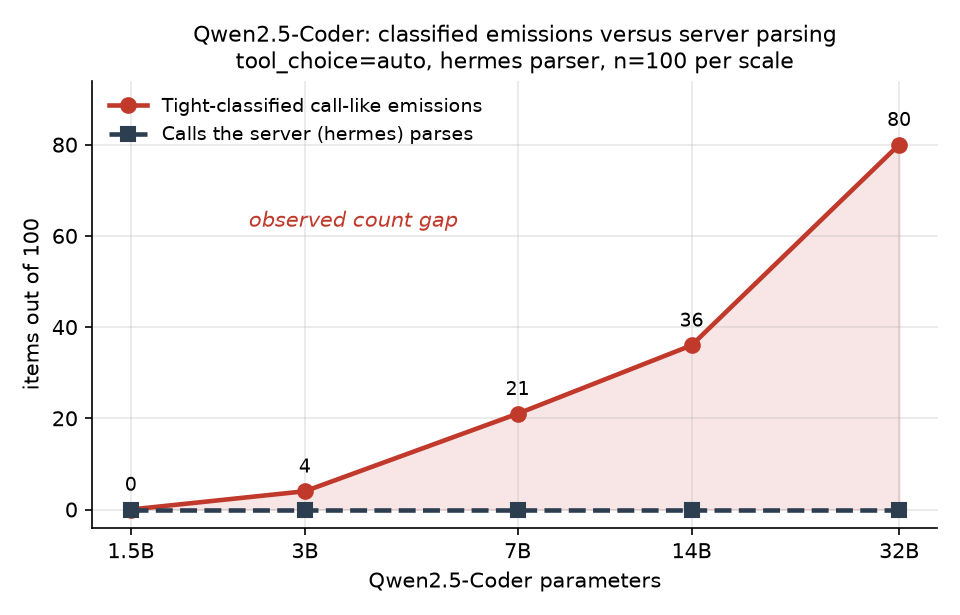}
\caption{Within-family observation under the mismatched interface. Across the five
Qwen2.5-Coder checkpoints, the server parses zero calls while tight-classified emissions
rise to 80/100 at 32B (approximately 72 as a sensitivity estimate using tight-stratum precision
0.900, assuming transfer across sizes; Appendix~\ref{app:humanval}). Same data as
Table~\ref{tab:scale}.}
\label{fig:gap}
\end{figure}

\subsection{Main table under unified configuration}
\label{sec:maintable_full}

\begin{center}
\begin{minipage}{\linewidth}
\centering
\small
\setlength{\tabcolsep}{4pt}
\captionof{table}{All arms genuinely at \texttt{max\_tokens}=2048, one shared 100-item set, \texttt{temperature}=0. Mistral has no admissible FC arm: all four configurations produce request errors, so no pass rate and no \emph{p}-value are computed.}\label{tab:main}
\setlength{\tabcolsep}{3.5pt}\begin{tabular}{@{}llllllll@{}}
\toprule
Model & \multicolumn{2}{c}{ReAct} & \multicolumn{2}{c}{FC} & b/c & \emph{p} & FC arm \\
\cmidrule(lr){2-3}\cmidrule(lr){4-5}
 & turn-1 & final & turn-1 & final & & & \\
\midrule
Llama-3.1-8B & 61 & \textbf{80} & 46 & 61 & 27/8 & 0.0019 & official tmpl.\ + \texttt{strict} \\
Qwen2.5-Coder-7B & 57 & \textbf{74} & 53 & 62 & 17/5 & 0.0169 & dedicated adapter \\
Mistral-7B-v0.3 & 26 & 33 & N/A & N/A & N/A & N/A & --- \\
\bottomrule
\end{tabular}
\end{minipage}
\end{center}

\subsection{The four Llama controls, tabulated}

\begin{center}
\begin{minipage}{\linewidth}
\centering
\small
\captionof{table}{Four single-variable controls on Llama-3.1-8B's 23\% wrong-tool rate. The first three did not reduce the error; the fourth eliminated it through a schema constraint, without identifying a unique cause of role confusion.}\label{tab:controls}
\begin{tabular}{@{}p{.27\linewidth}p{.21\linewidth}p{.27\linewidth}p{\dimexpr.25\linewidth-6\tabcolsep\relax}@{}}
\toprule
Alternative explanation & Control & Wrong-tool calls / 100 & Verdict \\
\midrule
schema lacks a parameter description & rich vs terse & 23 $\to$ 22 (\emph{p}=1.000) & rejected \\
ReAct merely adds a reasoning step & FC + Thought scaffold & 23 $\to$ \textbf{59} (\emph{p}\textless0.001, \textbf{worse}) & rejected \\
server not configured per documentation & + official chat template & 23 $\to$ 22 (1/0 discordance, \emph{p}=1.000) & rejected \\
\textbf{schema constraint not enabled} & \textbf{+ \texttt{strict: true} on official template} & \textbf{22 $\to$ 0 (22/0, \emph{p}=4.77$\times10^{-7}$)} & \textbf{accepted} \\
\bottomrule
\end{tabular}
\end{minipage}
\end{center}

\subsection{Conditioning on a successful parse, and access vs.\ use of feedback}
\label{app:conditioned}

\begin{center}
\begin{minipage}{\linewidth}
\centering
\small
\captionof{table}{Conditioning on items where both protocols parsed successfully. These are descriptive post-treatment subsets; neither nominal test clears the 27-test Bonferroni threshold.}\label{tab:conditioned}
\begin{tabular}{@{}llllll@{}}
\toprule
Family & Items both parsed & ReAct & FC & b/c & \emph{p} \\
\midrule
Qwen2.5-Coder-7B & 83 & 63 & 56 & 11/4 & 0.118 (n.s.) \\
Llama-3.1-8B & 96 & 78 & 60 & 25/7 & 0.0021 \\
\bottomrule
\end{tabular}
\end{minipage}
\end{center}

\begin{center}
\begin{minipage}{\linewidth}
\centering
\small
\captionof{table}{Descriptive repair success among items reaching a second code-bearing, tested step. The subset is selected after treatment; these rates do not localize or causally decompose the protocol gap.}\label{tab:given_feedback}
\begin{tabular}{@{}llll@{}}
\toprule
Arm & Reached tested step $\geq$2 & Rescued & Repair success given feedback \\
\midrule
ReAct & 40 & 17 & 42.5\% \\
FC + adapter & 37 & 9 & 24.3\% \\
FC + hermes & \textbf{0} & \textbf{0} & undefined \\
\bottomrule
\end{tabular}
\end{minipage}
\end{center}

\textbf{Adapters are not a universal switch.} The same adapter recovers 84/100 at 7B and
\textbf{1/100 at 3B}: the 3B checkpoint ignores the \texttt{<tools>} few-shot entirely (zero
\texttt{<tools>} tags in 100 items, 99 direct code) while still solving the task at a comparable
rate (final 53). Adapter effectiveness is non-monotonic in scale, reinforcing the per-checkpoint
claim of \S\ref{sec:families}. The dedicated solution also changes parser, chat template and
few-shot simultaneously; we additionally rewrote the template's few-shot examples, which
originally invoked \texttt{get\_weather} --- a tool absent from our schema, which would have
induced calls to a non-existent tool and inflated both initiation and empty-argument counts.
Original and modified hashes are recorded.

\subsection{Pressure on a broken interface makes things worse}
\label{sec:pressure}

Two independent controls point the same way. Adding a reasoning scaffold to FC raised role
confusion from 23 to \textbf{59} (\S\ref{sec:llama}). Replacing the optional tool instruction
with a mandatory one on Qwen2.5-Coder-1.5B moved parser-accepted calls not at all --- still
0/100 --- while final pass fell from \textbf{31 to 15} and unparsable outputs rose from
${\sim}0$ to 46/100. Neither intervention touches the interface, and both make the observable
outcome worse. \textbf{When an agent appears reluctant to use tools, strengthening the
instruction is the cheapest thing to try and the most likely to mislead} --- it can degrade task
performance while leaving the tool-call count at zero, which looks like a model that is both
unwilling \emph{and} incapable.

\begin{center}
\begin{minipage}{\linewidth}
\centering
\small
\captionof{table}{Probe replication on the 1.5B model actually being trained, recorded mandatory-prompt override (n=100; exact original file unavailable, Appendix~\ref{app:arms}). Not one output names \texttt{run\_tests}, so at this scale the missing calls are not a parser mismatch --- there was nothing well-formed to censor.}\label{tab:probe_categories}
\begin{tabular}{@{}lll@{}}
\toprule
Category & n & \% \\
\midrule
wrong name, \texttt{arguments} carry complete code & \textbf{0} & 0\% \\
wrong name, \texttt{arguments} are the task function's parameters & \textbf{36} & 69\% \\
wrong name, executable code present elsewhere in the body & 16 & 31\% \\
\bottomrule
\end{tabular}
\end{minipage}
\end{center}

\begin{center}
\begin{minipage}{\linewidth}
\centering
\small
\captionof{table}{Increasing pressure on the 1.5B model. Stronger instruction and few-shot examples raise the number of recoverable code payloads but never produce a call that names the tool.}\label{tab:pressure}
\begin{tabular}{@{}llll@{}}
\toprule
& recoverable calls & \textbf{naming \texttt{run\_tests}} & final pass \\
\midrule
mandatory prompt (baseline) & 52 & \textbf{0} & 15 \\
+ role-disambiguation few-shot & \textbf{64} & \textbf{0} & \textbf{13} \\
\bottomrule
\end{tabular}
\end{minipage}
\end{center}

\subsection{Formal 7B checkpoint curve and final-lineage audit}
\label{app:p3curve}

All checkpoint evaluations use the same 542-item multi-turn manifest, have zero request failures,
and attest the indicated native weight step. The accepted curve uses the original repaired steps
0--90 and resumed evaluation checkpoints 120 and 150; the abandoned original steps 91--116 are excluded. The repeated
step-90 evaluation after resumption gave 415/426/11 rather than 415/428/13 and is retained as
test--retest evidence, not substituted into the formal curve.

\begin{center}
\begin{minipage}{\linewidth}
\centering
\small
\captionof{table}{Complete formal 7B held-out curve. Each cell is turn-1 pass / final pass / rescued after turn 1, with $n=542$.}\label{tab:p3curve}
\begin{tabular}{@{}lrr@{}}
\toprule
step & broken FC & repaired FC \\
\midrule
0   & 418 / 430 / 12 & 418 / 430 / 12 \\
30  & 418 / 430 / 12 & 418 / 429 / 11 \\
60  & 420 / 433 / 13 & 417 / 431 / 14 \\
90  & 418 / 432 / 14 & 415 / 428 / 13 \\
120 & 419 / 433 / 14 & 415 / 427 / 12 \\
150 & 418 / 430 / 12 & 415 / 427 / 12 \\
\bottomrule
\end{tabular}
\end{minipage}
\end{center}

The policies did update. The accepted logs contain all 150 steps with non-zero actor gradient
norms: 0.0120--0.0251 for broken and 0.00641--0.01599 for repaired; mean training rewards span
0.150--0.469 and 0.168--0.435, respectively. A direct checkpoint audit over all 392 LoRA tensors
(80{,}740{,}352 parameters) finds 45{,}471{,}572 values changed from broken step 120 to 150
(56.3\%, $\ell_2$ delta 0.0787) and 45{,}296{,}056 from repaired step 90 to resumed step 150
(56.1\%, $\ell_2$ delta 0.1222). These comparisons establish updates to saved LoRA weights
over the retained intervals; evaluation completion records separately attest the native weight step.

The event join also resolves the funnel's apparent gaps. Broken FC has 19{,}203 generation events,
8{,}689 tight emissions and 10 accepted/executed/observed calls. Repaired FC has 33{,}367
generation events, 17{,}608 tight emissions, and 16{,}912 accepted calls in 16{,}844 accepted
generation events. Forty generations contain multiple calls, producing exactly the 68-call surplus;
the configured \texttt{max\_parallel\_calls=1} selects the first returned call and leaves
subsequent calls undispatched (neither merged nor queued by this loop). This accounts for the
68-call difference. All 16{,}844 dispatches join to executions and observations by request ID and assistant turn.
Extraction records lack those identifiers; accepted-generation and dispatch counts match per
trajectory, rather than through a per-generation join. Tight and accepted are different tests:
16{,}323 events satisfy both, 1{,}285 are tight but unaccepted, and 521 are accepted but not tight.
With the separately retained full archive (not included in the supplementary package),
\texttt{analysis/p3\_final\_evidence.py} verifies all evaluation and event hashes,
reconstructs the final lineage, and emits the per-file manifest; the server-side checkpoint audit is
reproducible with \texttt{analysis/p3\_checkpoint\_lora\_audit.py}.

\subsection{Historical ReAct positive control, tabulated}

\begin{figure}[htbp]
\centering
\includegraphics[width=\linewidth]{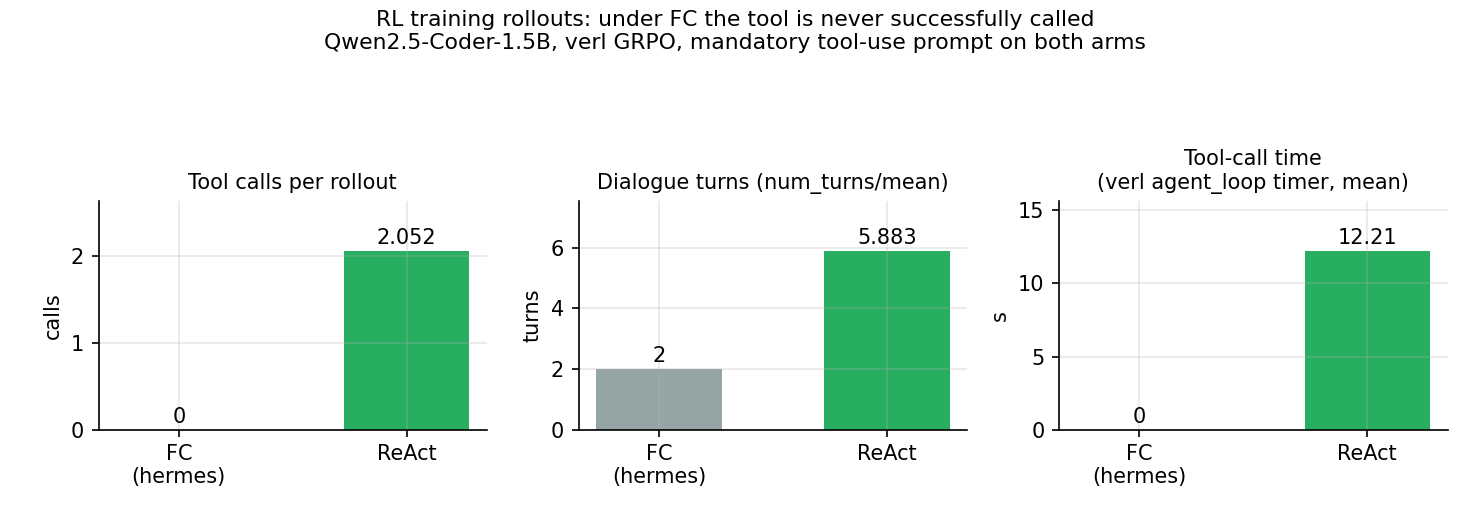}
\caption{\textbf{Historical 1.5B RL diagnostic.} Under function calling the tool is never
successfully called; \texttt{num\_turns} stays pinned at its minimum of 2 and tool time is exactly
zero. Prompt strength is matched across arms (both mandatory); step counts are not (10 vs 3).
The formal 7B parser control is reported separately in Table~\ref{tab:p3formal}.}
\label{fig:train}
\end{figure}

\begin{center}
\small
\begin{minipage}{\linewidth}
\centering
\captionof{table}{Historical ReAct training with a working tool channel, 23{,}676 executed calls. Rescues by turn $\geq$2 do not move. The three historical 1.5B FC runs, in which \textbf{zero} tool calls executed, give rescue sequences $[7,9,6,6,7,6,10,8,9,8,6]$, $[8,7,6,7,7,7,7,5,6,6,8]$ and $[7,5,7,5,7,8,9,9,8,9]$. This is supporting evidence only; Table~\ref{tab:p3formal} is the single-variable control.}\label{tab:sufficiency}
\begin{tabular}{@{}lrrrr@{}}
\toprule
step & turn-1 & final & \textbf{rescued by turn $\geq$2} & repair channel \\
\midrule
0  & 359 & 367 & 8 & 263 \\
15 & 340 & 346 & \textbf{6} & 266 \\
30 & 340 & 346 & \textbf{6} & 262 \\
45 & 341 & 349 & \textbf{8} & 261 \\
60 & 341 & 347 & \textbf{6} & 274 \\
\bottomrule
\end{tabular}
\end{minipage}
\end{center}

\subsection{Sampling variance, and an inverted noise structure}
\label{sec:variance}

Llama-3.1-8B, \texttt{temperature=0.6}, n=100 $\times$ 3 seeds.

\begin{center}
\begin{minipage}{\linewidth}
\centering
\small
\captionof{table}{Sampling variance. FC seed 3 contains one rejected parallel-call request. We report its complete-denominator endpoint, and treat the two clean arms separately; an SD from two clean seeds would be too unstable to summarize variation reliably.}\label{tab:variance}
\begin{tabular}{@{}lllll@{}}
\toprule
& seed 1 & seed 2 & seed 3 & admissible \\
\midrule
ReAct turn-1 & 53 & 56 & 43 & 3/3 \\
ReAct final & 72 & 72 & 72 & 3/3 \\
FC turn-1 & 44 & 44 & 45 & --- \\
FC final & 62 & 66 & 57$^{\dagger}$ & \textbf{2 clean / 3 end-to-end} \\
\bottomrule
\end{tabular}
\end{minipage}
\end{center}

FC seed 3 contains one request error --- the model emitted parallel tool calls and vLLM returned
\emph{``This model only supports single tool-calls at once''}. The end-to-end endpoint therefore
retains that task as a failure and is 57/100; the prespecified clean-arm analysis excludes the seed
and reports [62, 66]. An SD can be computed from two clean seeds, but would be an unstable estimate,
so we do not use it to characterize sampling variation. Worst case among clean arms: ReAct's
lowest (72) versus FC's highest (66) = \textbf{+6}.

\textbf{Inverted noise structure.} ReAct's turn-1 varies widely (43--56, sd 6.8) while its final
converges; FC's turn-1 is nearly constant (44--45, sd 0.58) while its final disperses. The
mechanism is compensation in the rescue count --- 19 / 16 / \textbf{29}, largest where turn-1 was
worst --- and in all three seeds the count of ``passed at turn 1 but failed finally'' is
\textbf{0}: the repair loop is monotone. \textbf{The three identical finals are a coincidence of
totals, not identical item sets}, and sd = 0.00 must not be read as ``ReAct is perfectly
stable''.

\subsection{Exploratory observations of stable totals and changing item sets}
\label{sec:composition}

Two independent observations show the same structure. \emph{Across sampling seeds}: ReAct's final
pass is 72 / 72 / 72 at \texttt{temperature=0.6}, yet the solved sets have pairwise Jaccard
0.71--0.80 --- 15--24 items differ. \emph{Across training steps}: between steps 15 and 30 of the
ReAct run, 85 of 542 first-turn completions changed verbatim and 6 items flipped outcome (3
gained, 3 lost), leaving turn-1 and final pass numerically identical (340 / 346 at both
checkpoints). The policy is demonstrably moving; the aggregate is not.

These exploratory observations show that equal aggregate scores can conceal different
item-level outcomes. Compensation through rescues is consistent with the observed counts,
but these few runs do not identify a general stabilising effect of repair or establish a
deployment benefit. We therefore report item-level changes alongside totals; both observations
come from one model family and one task.

\subsection{The training verifier: seams, audit, and planned fixes}
\label{app:verifier}

The GRPO reward is outcome-only: it extracts the last submitted program, runs the item's own
\texttt{pytest} suite in the sandbox, and returns the pass ratio (\texttt{reward\_code.py}). Three
reward-hacking routes are closed by construction --- a run in which no test executes scores 0
rather than 1, a timeout scores 0, and an item with no test string scores 0. The historical
verifier nevertheless had two seams. \emph{(i)} Its summary parser searched
\texttt{(\textbackslash d+) passed} over merged stdout/stderr and took the first match, so model
stdout could imitate pytest's summary. \emph{(ii)} \texttt{skipped} and \texttt{xfail} were not
counted in the denominator. Both were reachable in principle. We therefore audited:
across the five training logs that retain generated code there are \textbf{zero} occurrences of a
printed \texttt{``N passed''}, of \texttt{pytest.skip}/\texttt{allow\_module\_level}, or of
\texttt{sys.exit}/\texttt{os.\_exit}, and \texttt{critic/rewards/mean} oscillates in 0.46--0.73
without the step change a discovered exploit produces. The audit's own limit is that historical
full rollout text was not retained, so it covers only the fragments those logs carry. Before the
formal 7B control we anchored the regex to pytest's summary, included
\texttt{skipped}/\texttt{xfailed}/\texttt{xpassed} in the denominator, and retained every rollout
in full with a SHA-256 digest. Both P3 arms use this fixed verifier. Neither verifier path is used
for the reported held-out pass rates, which require \texttt{mode="script"}, a zero exit code and
\texttt{status=="ok"}.

\subsection{Post-hoc common native function-calling evaluation}
\label{app:nativefc}

This additional evaluation uses the existing base and final-lineage step-150
broken/repaired checkpoints, not new training. It was designed after review, outside
the training preregistration. All three weights see the same 542 frozen EvalPlus
items, original tokenizer/chat template and \texttt{run\_tests} schema, and the same
working repaired verl extraction logic. The BFCL few-shot adapter is not used.
Each loaded LoRA has 392 tensors and 80,740,352 parameters; ordered float32 tensor
digests match the audited checkpoint exports, with rank/alpha 32 and scaling 1.
Base/tokenizer configuration hashes are checked; full base-weight shards were not
independently rehashed. Generation uses sequential Transformers with bf16/SDPA,
greedy decoding, seed 0, at most four assistant generations of 1,024 new tokens,
an 8,192-token context budget and a 1,200-character observation cap. Context overflow
fails explicitly, without silent truncation. The backend is PyTorch 2.13.0, Transformers 5.10.4, PEFT 0.20.0 and safetensors 0.8.0.
Configuration records, including schema, task IDs and source hashes, are in the public
repository (\texttt{submission\_packages/native\_fc/configurations.json}).

Only an accepted, correctly named call with nonempty code triggers interactive
execution. The first accepted call per generation is dispatched; additional calls
would be ignored, but none occur here. Feedback is a tool-role message and full
history is re-rendered on every turn. Passing a tool test alone does not stop the
loop; a no-call response or the generation cap ends it. A later explicit final
code submission supersedes the last tool code; plain closing prose does not.
After the dialogue, final code is scored once independently with no returned
feedback. That terminal test is not a tool invocation. Tests use EvalPlus script
mode with a 30-second timeout, not the training reward's pytest partial credit.
This aligns tool-role feedback, but is not byte-identical replay of AgentLoop or
an elimination of task/test distribution shift.

\begin{table}[htbp]
\centering
\caption{Common native FC evaluation, 542 paired tasks per weight. Calls are
interactive tool executions; timeouts are terminal scoring timeouts.}
\label{tab:nativefc}
\small
\begin{tabular}{lrrrrr}
\toprule
Weight & Final pass & Initiating tasks & Calls & Tool rescues & Timeouts \\
\midrule
Base & 239 & 46 & 91 & 2 & 4 \\
Broken step 150 & 248 & 50 & 93 & 1 & 5 \\
Repaired step 150 & 248 & 47 & 90 & 1 & 6 \\
\bottomrule
\end{tabular}
\end{table}

Here a rescue requires a failed first \emph{actual tool test} followed by a passing
tool test. An untested first response is not a failure rescued later. Thus these
counts are not interchangeable with the earlier automatic-feedback rescue metric.
A complementary descriptive count asks whether a failed first tool test is followed
by a passing final submission, including a submission not retested interactively:
this gives 2/2/1 for base/broken/repaired. Broken MBPP/612 fails three tool tests,
then submits passing final code without another tool call. The stricter tool-test
rescue count of one therefore does not mean that only one final submission was
repaired after feedback. Neither definition shows more rescues for repaired FC.
Initiation here means at least one accepted and executed call, not every attempted
or call-shaped emission.
Accepted calls, executions and returned observations agree (91/93/90); generation
counts are 619/622/619. No output hits the generation length cap. Every interactive
execution completes with status \texttt{ok}; this denotes successful test execution,
not necessarily code correctness. Terminal timeouts remain failures in the full
542-item denominator, with no clean-arm filtering. Base timeouts are MBPP 84, 126,
388 and 739; broken additionally times out on 603; repaired additionally on 603
and 765. These observed timeouts are not independently diagnosed as intrinsic
nontermination. No missing submissions or other terminal infrastructure statuses
are recorded.

The historical programmatic-feedback evaluator records failures on both previously
flagged items (HumanEval/32 and MBPP/599); this does not transfer automatically to
the new protocol. In native FC, HumanEval/32 fails for all weights, whereas MBPP/599
fails for base but passes for both trained weights. We preserve the frozen 542-item
primary analysis. Excluding both historically flagged items gives final success
239/540, 247/540 and 247/540. The trained-arm discordances remain 16/16; the base
contrasts lose one trained-only success. This descriptive sensitivity does not
reclassify either test as defect-free or diagnose the reason for passing it.

Absolute success also differs substantially across protocols: the trained weights
score 430/542 and 427/542 under programmatic feedback versus 248/542 each under
native FC. This is not an isolated estimate of the benefit of automatic feedback:
the backend, history rendering, code-submission handling and stopping rules also
change. We interpret the within-protocol weight contrasts, not the cross-protocol
gap as a causal effect. Reduced records cannot independently audit those controller
choices; full raw trajectories remain available in the retained research archive.

For repaired minus broken final success, gains/losses are 16/16, difference zero
percentage points, conservative paired 95\% CI $[-3.47,3.47]$, exact two-sided
McNemar $p=1$. Relative to base, broken has 20 gains and 11 losses
($+1.66$ points, CI $[-1.74,5.03]$, $p=0.150$), and repaired 22 gains and 13 losses
($+1.66$ points, CI $[-1.94,5.23]$, $p=0.175$). These intervals use the construction
in Appendix~\ref{app:statistics}; they cover item-pair uncertainty, not training-seed
or inference-retest variability. Equal totals do not imply equal passing sets or
equivalent policies. Neither protocol detects a success benefit of repair in this
setting; this does not establish that interface repair cannot improve learning.

The three weight contrasts crossed with final success, initiation and rescue form
nine explicitly post-hoc exploratory tests, with Bonferroni-adjusted values supplied
alongside nominal values. They are separate from the earlier 27-test reporting
audit; none is significant even before adjustment. A single 36-test correction
would not make any of the new results significant. The intervals are pointwise,
not simultaneous. The deterministic six- and 24-task engineering pilots overlap
the full set and are not counted as additional observations or seeds.

The public repository supplies these configurations, the 1,626 item records
(\texttt{submission\_packages/native\_fc/}) and
\texttt{native\_\allowbreak fc\_\allowbreak results.py}. Running this CPU script with \texttt{--outcomes}
recomputes every listed contrast and timeout inventory without GPU access.
The reduced records do not contain generated code or full conversations;
independent reclassification or execution requires the separately retained raw
archive. Its \texttt{--archive} mode checks raw hashes, matching inputs,
loaded adapter identity and execution-to-observation links before reduction.

\section{Limitations, in full}
\label{app:limitations}

In the order a sceptical reader should apply them.

\textbf{Scope of the evidence.}
\emph{(1) One task family for the scale and training results.} Our own probe uses decontaminated
KodCode items with a single tool taking one \texttt{code} argument. The external-validity results
are \emph{not} code-only: BFCL's \texttt{multi\_turn\_base} spans file-system, vehicle and trading
APIs, and $\tau$-bench \texttt{retail} is a customer-service domain whose 789 unparsed turns
decompose to 100\% envelope-layer with zero parser loss, reproducing the code result
(\S\ref{sec:tau}). What remains code-only is the \emph{scale ladder} and the \emph{RL
contamination} result. The former is a real gap --- the same ladder on a non-code task would test
whether the undercount's growth with checkpoint size is a property of the interface or of
code-shaped payloads (\S\ref{sec:future}). The latter we do not expect to close: the reward is a
\texttt{pytest} pass ratio, so porting it changes environment, reward and task at once. The controlled
scale and RL experiments use a single-tool schema and do not isolate tool selection among
alternatives. A single-tool schema is
an easy parser case, but we treat that only as motivation for broader tests, not as a quantitative
lower-bound argument.
\emph{(2) Item sampling.} The 50 full-length arms use 100 items taken as \texttt{clean[:100]},
\textbf{not a random sample}; pairing across arms is exact, so within-study comparisons are sound.
Generalisation rests on a replication on a random 300-item sample (seed 20260901): server-parsed
remains 0 at every size across 1500 items, and strict emitted counts scale to 0, 4, 30, 40, 81 per
100 against the original 0, 4, 21, 36, 80. \textbf{7B moves from 21 to 30}, so the original subset
understated that rung; direction and monotonicity are unaffected, but a single 100-item subset is
evidently not a stable estimate of an intermediate rung.
\emph{(3) The scale result is within one family.} The $0\to4\to21\to36\to80$ ladder is
Qwen2.5-Coder alone, against one mismatched interface, on one task; we do not claim censoring is
generally scale-increasing, and we do not equate parameter count with capability. The Instruct
ladder is a control on the \emph{mechanism}, not a second ladder of undercount --- its execution
layer was initially void (the host lacked \texttt{pytest}) and has since been re-run with a
working executor, the parse layer reproducing value for value. A second \emph{family}'s ladder
remains the test that would promote or refute the scale claim.
\emph{(4) The taxonomy is four instances, not an enumeration.} The evidence is deliberately
asymmetric: Qwen has a scale ladder, a replay matrix and a repair; Llama a single-variable
intervention; \textbf{Mistral has no admissible 100-item FC comparison at all}; DeepSeek is a
template check with little quantification.
\emph{(5) Single-sample headline arms.} Table~\ref{tab:repair} and the ladders are one sample at
\texttt{temperature}=0. Variance is estimated only on Llama-3.1-8B, and there only 2 of 3 FC arms
are clean. A standard deviation from two seeds is defined but too unstable to summarize sampling
variation reliably; the complete-denominator endpoint for the error-bearing seed is still
reported (\S\ref{sec:variance}).
\emph{(6) Multiple comparisons.} The explicit 27-test retrospective family in Appendix~\ref{app:statistics} uses
$0.05/27\approx0.00185$. This reporting rule was not preregistered. Surviving it: the official-template \texttt{strict: true}
intervention on wrong-tool labels ($p=4.77\times10^{-7}$) and final pass
($p=7.63\times10^{-5}$), the protocol gap on 300 random items ($p$=3.1$\times$10$^{-6}$), and the official-template strict turn-1 contrast ($p=0.000977$). \textbf{Not surviving it}: the Llama strict-to-ReAct contrast ($p=0.0019$),
the both-parsed Llama contrast ($p=0.0021$), the repair-loop pass gain ($p$=0.093 at
$n$=100, $p$=0.0385 at $n$=300), the Llama protocol contrast at $n$=300 ($p$=0.0352), and the
residual-gap estimates ($p$=0.118, 0.125). We rest no claim on the latter group, and report the
$n$=300 values precisely so a reader can see the direction is stable where significance is not
claimed.
\emph{(7) Model scale stops at 32B}, on a single A800; no frontier models.

\textbf{Threats to the causal reading.}
\emph{(8) The clean repaired-FC control has one seed and one budget.} The completed 7B experiment
holds model, LoRA configuration, data, prompt, decoding, verifier, seed and 150-step budget fixed,
changing only the parser registered in verl. It therefore identifies the interface effect on the
sampled experience distribution in this setting. Its null held-out result does not establish that
a working channel is generally insufficient, and its LoRA training is not strictly comparable to
the historical 1.5B full-parameter runs.
\emph{(9) Conditioning on a successful parse is post-treatment conditioning.} Whether an item
parses is determined by the interface under test, so the both-parsed subset is not a random
subsample and its contrast is not a clean causal estimate. On Qwen the residual is $+8.4$\,pp,
95\% CI $[-0.5,+17.4]$, $p$=0.118: \textbf{we do not detect a residual difference, which is not
the same as establishing there is none}.
\emph{(10) Near-zero observed frequency is not zero probability.} The broken 7B arm executes 10
calls rather than none; its instrumentation describes a sampled distribution, not the support of
the policy. We therefore report a nearly closed channel (10 executions alongside 8{,}689 tight-classified
generation events), not a mathematically unreachable action.
\emph{(11) Attempts and discards are separated only in the new run.} The formal 7B arms retain
complete rollout text with hashes, parser verdicts, execution events and returned observations, so
8{,}689 regex-positive emissions and 10 accepted actions are separately recorded; the classifier
is not a semantic-validity oracle and its link to dispatch is a trajectory-level count check. The historical 1.5B
run did not retain full text. Its same-prompt replication finds no correctly named call, but that
cannot retroactively identify what every historical rollout emitted.
\emph{(12) The null learning observation is setting-specific.} Parser repair restores 16{,}844 executed
and observed calls over 150 steps with 12 held-out rescues at both endpoints (intermediate checkpoints vary). This is the requested
single-variable control, but it is one seed, 7B, LoRA, one code task and one training horizon.
The earlier ReAct arm (23{,}676 calls, flat rescues) is directionally consistent but remains
confounded by its protocol change.

The shared evaluation deliberately avoids both training parsers, but changes structured tool-role
feedback to user-message feedback. It measures transfer to this programmatic-feedback protocol,
not native FC behavior. Formatting shift, limited adaptation budget/capacity and multi-turn
credit assignment are plausible explanations for absent transfer; these data do not separate them.
The evaluator extracts the last decodable \texttt{<tool\_call>} code argument when present,
otherwise the last code fence or the raw text. This is not a general decoder for every tool syntax.
The post-hoc common native FC evaluation (Appendix~\ref{app:nativefc}) additionally
uses voluntary calls and tool-role feedback: repaired and broken both pass 248/542,
with 16 discordances each way. This addresses the missing initiation measurement,
not all distribution shift: held-out tasks use script-mode tests, the controller
re-renders history, and only one training seed is available. It is not a byte-identical
training rollout or an equivalence test. The native FC tool-test rescue definition requires an actual
failed tool test followed by a passing tool test; a final-submission alternative is
reported separately. Neither may be pooled with programmatic-feedback rescues.
Absolute scores differ across evaluation protocols, so their cross-protocol gap
must not be attributed solely to whether testing is automatically initiated.

\emph{(13) Human validation covers one criterion on one sample.} The 98-item stratified sample has 97 complete paired annotator records; these give inter-annotator $\kappa$ = 0.967 on the 62 items where both read identical
bytes, and classifier-vs-gold-standard $\kappa$ = 0.871 after third-party adjudication
(Appendix~\ref{app:humanval}). Two residual limits: the sample is stratified by classifier
verdict, so the unweighted sample metrics are not population precision or recall; and stored raw
output is capped at 4000 characters, so a call emitted past that point is invisible to classifier
and annotators alike, creating false negatives. Because the classifier also has false positives,
the net bias of the raw emitted-call count is not identified. The \emph{first} round's apparent
unreliability ($\kappa$=0.713) was confounded by 2400-character truncation in the annotation
pack and by selective adjudication; the identical-byte analysis removes the presentation confound
but does not establish that every disagreement was an instrument error.
\emph{(14) The historical training verifier had two exploitable seams.} Its summary parser used
the first unanchored \texttt{(\textbackslash d+) passed} match and omitted
\texttt{skipped}/\texttt{xfail} from the denominator. We found no exploit signature in the five
historical logs that retain generated code, but those logs do not contain every full rollout. For
the formal 7B control we anchored the regex to pytest's summary, included skipped, xfailed and
xpassed outcomes in the denominator, and retained every rollout in full with a hash. Both arms use
the identical fixed verifier, so it cannot explain their contrast. Neither verifier version is
used for the separately reported held-out pass rates (\S\ref{app:verifier}).

\textbf{Environment dependence and provenance.}
\emph{(15) Serving-stack version dependence.} Serving and training measurements use vLLM 0.27.1 and verl
0.9.0, so parser behaviour, template defaults and the semantics of \texttt{strict: true} are
properties of those versions. We pin versions, parser and template hashes, and model revisions for
exactly this reason, and we have not replicated on a second serving stack (e.g.\ SGLang), so we
cannot say which findings are vLLM-specific. The post-hoc native FC evaluation uses
Transformers 5.10.4/PEFT 0.20.0 rather than a second serving-stack replication;
its controller and test protocol differ from training and the earlier evaluator.
\emph{(16) Repository-revision dependence.} Chat templates ship inside
\texttt{tokenizer\_config.json} and can be changed by the publisher without a model change; the
Qwen2.5-Coder false positive is precisely an artefact of an inherited template.
\emph{(17) Instrumentation errata, disclosed rather than absorbed.} Seven ReAct arms recorded
\texttt{max\_tokens=2048} while running at 1024 (an asymmetry that \emph{favoured} FC; re-running
changed final pass by $\leq$1 item); Llama's initiation rate required manual correction from 97 to
74 after tool-name validation was added; five arms are error-bearing and admissible only for the
error-rate census, and one previously reported $p$=0.648 computed on such an arm is withdrawn. All
are enumerated in Appendix~\ref{app:errata}. None is silently absorbed into a reported number.

\subsection*{What would change our conclusions (restated)}

We name these so the claims are falsifiable rather than merely hedged.
\emph{Replications or a longer repaired-FC budget.} The requested broken-versus-repaired
single-variable control is now complete: parser repair restores 16{,}844 tool interactions and
does not increase held-out rescues. A consistent rescue gain over additional seeds or a longer
preregistered horizon would narrow the scope of this 150-step, single-seed observation; continued
nulls would strengthen it within, but not beyond, the tested model, task and optimiser setting.
\emph{A second family's scale ladder}, with a working test executor: if the undercount does not
grow with checkpoint size in another family, \S\ref{sec:scale}'s scale claim is a
Qwen-plus-\texttt{hermes} fact rather than a scaling phenomenon.
\emph{The same ladder on a non-code task.} The five Qwen2.5-Coder checkpoints run against
$\tau$-bench \texttt{retail} under the documented configuration --- models and interface fixed,
only the domain varied. If the emitted column still rises with size while the parsed column stays
at zero, the scale result is a property of the interface; if it does not, it is a property of
code-shaped payloads. This needs a rescue arm at the small sizes for the reason Granite did: a
checkpoint that simply never attempts a call produces an uninformative zero, not a censored one.
\emph{The same five-layer measurement on further standard benchmarks.} Done for BFCL~v4 and for
$\tau$-bench \texttt{retail} (\S\ref{sec:bfcl}), so the measurement-validity concern is
\emph{not} specific to our harness, and it now covers one interactive suite with a simulated user
and stateful environment. What the $\tau$-bench arm does \emph{not} establish is an outcome
effect: 7 versus 10 tasks solved on three discordant pairs is far from significant, and its user
simulator is a local Llama-3.1-8B rather than \texttt{gpt-4o}, so its absolute scores are not
comparable to the published leaderboard. Suites whose tool schemas are much larger, or whose
scoring depends on argument-level matching, remain untested.

\section{Statistical estimands and retrospective test family}
\label{app:statistics}

\subsection{Finite-sample interval for the formal RL endpoint}
Let $p_+$ denote the probability that an evaluation item passes only under repaired FC and
$p_-$ the probability that it passes only under broken FC. The paired risk difference is
$\Delta=p_+-p_-$. At step 150, the counts are $(n_+,n_-,n)=(0,3,542)$.
For each discordance probability we invert the binomial distribution to obtain a two-sided
97.5\% Clopper--Pearson interval, denoted $[L_+,U_+]$ and $[L_-,U_-]$.
The union bound gives at least 95\% coverage for
$[L_+-U_-,\,U_+-L_-]$ under independent evaluation-item pairs. This conservative construction
does not rely on a normal approximation with three discordant pairs. The point difference is
$-0.55$ percentage points and the interval is $[-1.79,0.72]$ percentage points.
For repaired step 0 versus 150, $(1,4,542)$ gives $[-2.06,1.01]$ percentage points.
These are unadjusted endpoint intervals, not simultaneous intervals for the whole test family.
They quantify item-pair uncertainty under an independence assumption, not training-seed,
inference-retest, benchmark-selection or task-cluster uncertainty.

\subsection{Explicit reporting family}
This is a retrospective reporting audit, not a preregistered multiplicity policy.
The family includes all 15 metrics in the five audited Llama control contrasts, even when only
one metric is highlighted, plus the 12 distinct comparisons below. Repeated appearances of the
same contrast count once. Nominal $p$ values are exact two-sided McNemar tests;
adjusted $p=\min(1,27p)$, with threshold $0.05/27=0.00185185$.
Historical 1.5B motivating runs and abandoned/invalid runs are not confirmatory evidence and
are outside this family; their original disclosures remain in the historical audit.
The two both-parsed contrasts are retained in the family but remain post-treatment descriptive
comparisons, not causal effects. This explicit family replaces the earlier approximate count of
twelve; in particular, the Llama strict-to-ReAct and both-parsed contrasts no longer clear
the corrected threshold.

\begin{center}
\scriptsize
\begin{longtable}{@{}lp{.49\linewidth}rrr@{}}
\caption{Retrospective test inventory. G/L are later-arm gains/losses in the named binary metric;
for wrong-tool metrics a gain means more wrong-tool calls, not an improvement. A star marks
$p\leq0.05/27$.}\label{tab:testfamily}\\
\toprule
ID & Contrast and metric & $n$ & G/L & nominal $p$\\
\midrule
\endfirsthead
\toprule ID & Contrast and metric & $n$ & G/L & nominal $p$\\ \midrule
\endhead
T01 & Llama terse to rich: wrong\_tool & 100 & 5/6 & 1.00 \\
T02 & Llama terse to rich: turn1\_pass & 100 & 4/6 & 0.754 \\
T03 & Llama terse to rich: final\_pass & 100 & 10/10 & 1.00 \\
T04 & Llama terse to thought: wrong\_tool & 100 & 37/1 & 2.84e-10* \\
T05 & Llama terse to thought: turn1\_pass & 100 & 1/23 & 0.00000298* \\
T06 & Llama terse to thought: final\_pass & 100 & 4/29 & 0.0000109* \\
T07 & Llama terse to official: wrong\_tool & 100 & 0/1 & 1.00 \\
T08 & Llama terse to official: turn1\_pass & 100 & 0/1 & 1.00 \\
T09 & Llama terse to official: final\_pass & 100 & 1/6 & 0.125 \\
T10 & Llama terse to strict: wrong\_tool & 100 & 0/23 & 2.38e-7* \\
T11 & Llama terse to strict: turn1\_pass & 100 & 13/1 & 0.00183* \\
T12 & Llama terse to strict: final\_pass & 100 & 15/3 & 0.00754 \\
T13 & Llama official to strict: wrong\_tool & 100 & 0/22 & 4.77e-7* \\
T14 & Llama official to strict: turn1\_pass & 100 & 14/1 & 0.000977* \\
T15 & Llama official to strict: final\_pass & 100 & 18/1 & 0.0000763* \\
T16 & Llama strict to ReAct final & 100 & 27/8 & 0.00188 \\
T17 & Llama terse to ReAct final & 100 & 38/7 & 0.00000312* \\
T18 & Qwen Hermes to adapter final & 100 & 16/7 & 0.0931 \\
T19 & Qwen adapter to ReAct final & 100 & 17/5 & 0.0169 \\
T20 & Qwen Hermes to adapter final (random) & 300 & 43/25 & 0.0385 \\
T21 & Qwen adapter to ReAct final (random) & 300 & 59/18 & 0.00000306* \\
T22 & Llama strict to ReAct final (random) & 300 & 51/31 & 0.0352 \\
T23 & Qwen both-parsed FC to ReAct final & 83 & 11/4 & 0.118 \\
T24 & Llama both-parsed FC to ReAct final & 96 & 25/7 & 0.00210 \\
T25 & tau documented to repaired success & 115 & 3/0 & 0.250 \\
T26 & P3 broken to repaired final at step150 & 542 & 0/3 & 0.250 \\
T27 & P3 repaired step0 to step150 final & 542 & 1/4 & 0.375 \\
\bottomrule
\end{longtable}
\end{center}
The accompanying \texttt{submission\_statistics.py} recomputes the 15 Llama tests from
item-level records and the remaining tests from the explicit paired counts, with their sources
identified in \texttt{submission\_statistics.json}. Thus nominal and adjusted values can be
reproduced without GPUs. The latter count-based recomputation is not a new item-level audit
of every historical comparison.

The later post-hoc native FC evaluation adds a separately identified nine-test
exploratory family (Appendix~\ref{app:nativefc}): three paired weight contrasts
crossed with final success, tool initiation and rescue. Its per-item inputs,
nominal and Bonferroni-adjusted values are supplied separately. No new contrast
is nominally significant, and none becomes significant if all 36 tests are pooled.
The historical 27-test audit is preserved, not retroactively called preregistered.

\end{document}